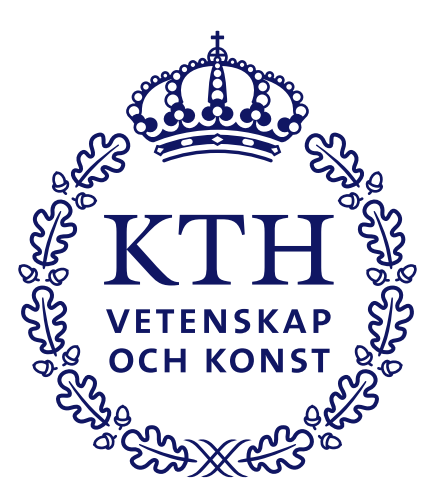

Doctoral Thesis in Computer Science

# Variational Mixtures and Multi-Marginal Flow Matching

## Advancing Statistical Inference with Biological Applications

OSKAR KVIMAN

KTH ROYAL INSTITUTE OF TECHNOLOGY

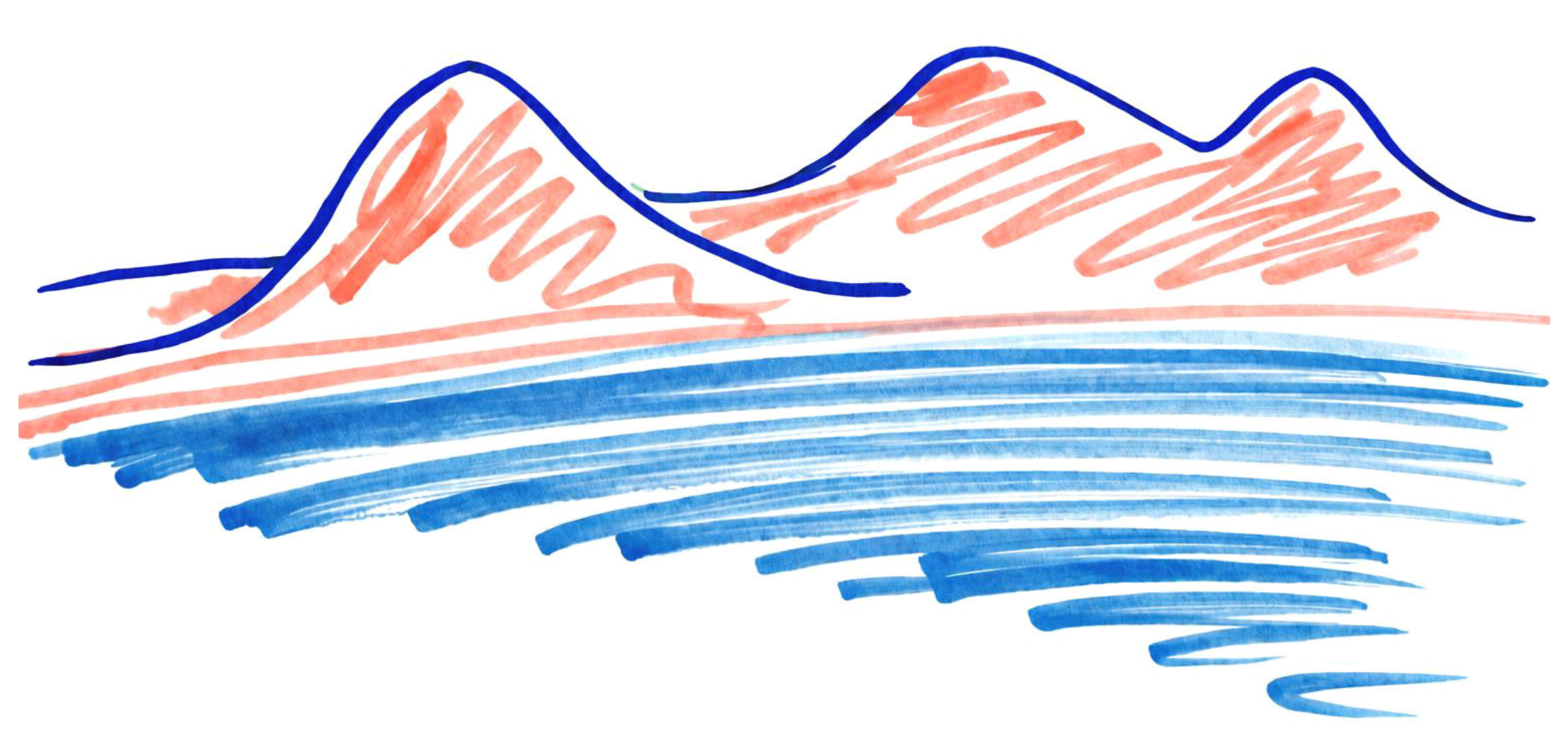

# Variational Mixtures and Multi-Marginal Flow Matching

## Advancing Statistical Inference with Biological Applications

OSKAR KVIMAN

Cover page photo: Oskar Kviman

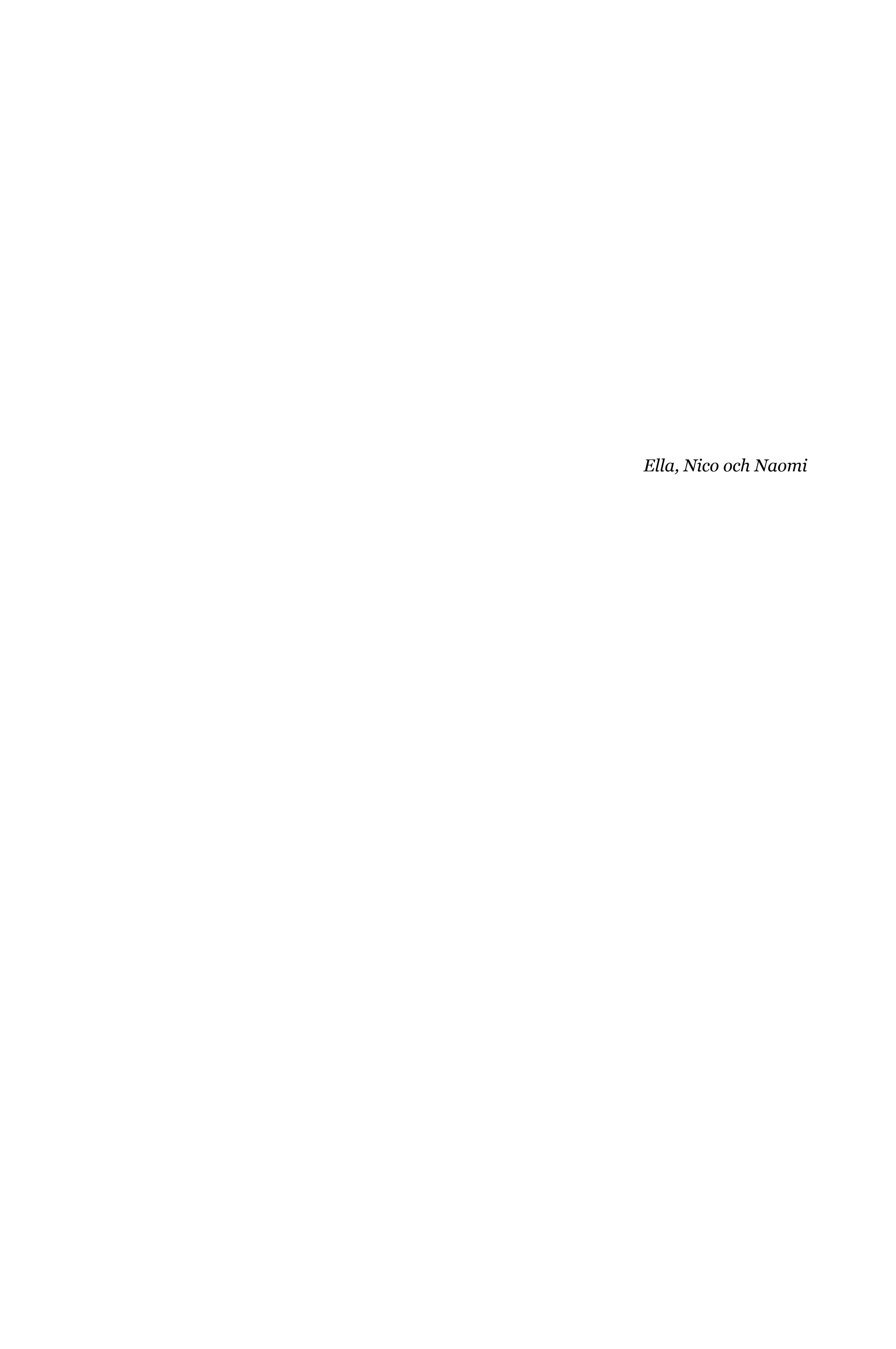

*Ella, Nico och Naomi*

The theory of probabilities is at bottom nothing but common sense reduced to calculus

---

*Pierre-Simon Laplace*

# Abstract

In this thesis I develop methods for statistical inference when the distributions arising from complex biological systems are multi-modal, geometrically structured, and sometimes only defined up to a normalizing constant. I start from variational inference and, when analytic update equations are unavailable, move to black-box variational inference. To build intuition regarding inference challenges and the proposed methodologies, I introduce a novel unnormalized target density (the CoLN distribution) and reuse it as a controlled test case in the kappa. I then trace a trajectory of increasingly expressive approximations: ensembles evaluated with the multiple importance sampling ELBO (Paper A) and variational mixtures that automate component cooperation and exploration (Paper B). Because expressivity comes at a cost, I develop efficient mixture learning ideas, including Monte Carlo objective estimators to scale mixture learning more efficiently (Paper C). As a new result in the kappa, I overturn a three decades long misconception regarding the potential performance benefits of using mixtures in variational inference. Finally, I move from variational inference to flow matching, where I address the need for specialized treatment of interpolant learning in multi-marginal settings (Paper D). By combining insights from Papers A–D, I derive in Section 5.5 a new method: multi-marginal flow matching with mixtures of variational interpolants. I connect these methodological developments to biological applications, with special emphasis on three-dimensional spatial transcriptomics, where stacked tissue slices induce multi-modal dynamics across space.

**Keywords**
variational inference, black-box variational inference, variational mixtures, multi-marginal flow matching, spatial transcriptomics

# Sammanfattning

I denna avhandling utvecklar jag metoder för statistisk inferens när de fördelningar som uppstår i komplexa biologiska system är multimodala, har geometriska strukturer och som ibland endast är definierade upp till en normaliseringskonstant. Jag utgår från variationsinferens och går, när analytiska uppdateringsekvationer saknas, vidare till black-box variationsinferens. För att göra utmaningen konkret introducerar jag en onormaliserad måldistribution (CoLN-fördelningen) och återanvänder den som ett kontrollerat testfall i kappan. Därefter följer jag en bana av allt mer uttrycksfulla approximationer: ensembler utvärderade med multiple importance sampling ELBO (MISELBO) (Artikel A) och variationsblandningar som automatiserar utforskning och samverkan mellan komponenter (Artikel B). Eftersom uttrycksfullhet kommer till följd av ökade beräkningskostnader utvecklar jag idéer för effektiv blandningsinlärning, inklusive ett amortiseringsschema för att skala blandningsinlärning mer effektivt (Artikel C). Slutligen går jag från variationsinferens till flow matching, där ordinära differentialekvationer definierar fördelningsdynamik, och behandlar behovet av specialiserade metoder i multi-marginala problem (Artikel D). Genom att kombinera insikter från Artiklar A–D härleder jag i Avsnitt 5.5 en ny metod: *multi-marginal flow matching with mixtures of variational interpolants*. Jag kopplar dessa metodutvecklingar till biologiska tillämpningar, med särskild betoning på tredimensionell spatial transkriptomik, där staplade vävnadssnitt ger upphov till multimodal dynamik genom rummet.

# Acknowledgment

Jens Lagergren, your generosity with time, advice and trust is remarkable and has been the fundamental ingredient for the success of this thesis. Beyond problem solving skills and oenological expertise, I am happy to have gained a friend—thanks for these fantastic years!

Víctor Elvira, you have acted as a co-supervisor to me, and I am deeply grateful to you for sharing your brilliance on statistics and method development. I cherish our friendship and I look forward working more together in the future.

To all the past and current fellow lab members, thank you for your support and your positive influence on my work. Special thanks to my co-authors Alexandra Hotti, Ricky Molén, Hazal Koptagel, Hosein Toosi, Semih Kurt, Negar Safinianaini, Pedro Ferreira and my dear friend Harald Melin.

I have had the privilege to work with outstanding researchers outside of our lab, too. Among many others, I want to acknowledge the contributions of Nikolay Malkin, Nicola Branchini and Kirill Tamogashev as particularly important for the success of this thesis.

My Ph.D. journey would have been much less fulfilling without my supportive and curious friends and relatives, and impossible without my ancestors who paved the way for me to reach this point.

Paula, mamma och pappa, thanks for your encouragements and endless confidence boosting support.

Nico och Naomi, you have taught me more than any doctoral education ever could, and you have made life more meaningful than any imaginable degree. This thesis is dedicated to you, and to Ella: my best friend. Ella, thank you for your loving patience and for being the best life companion. Your impact on this thesis is beyond words.

# List of included papers

The list of papers by the author included in the doctoral dissertation.

**Paper A** **Multiple importance sampling ELBO and deep ensembles of variational approximations**, Oskar Kviman, Harald Melin, Hazal Koptagel, Victor Elvira, and Jens Lagergren. In International Conference on Artificial Intelligence and Statistics, 2022 [1]

**Paper B** **Cooperation in the latent space: The benefits of adding mixture components in variational autoencoders**, Oskar Kviman, Ricky Molén, Alexandra Hotti, Semih Kurt, Vıctor Elvira, and Jens Lagergren. In International Conference on Machine Learning, 2023 [2]

**Paper C** **Efficient mixture learning in black-box variational inference**, Alexandra Hotti[1], Oskar Kviman[1], Ricky Molén, Víctor Elvira, and Jens Lagergren. In International Conference on Machine Learning, 2024 [3]

**Paper D** **Multi-marginal flow matching with adversarially learnt interpolants**, Oskar Kviman[1], Kirill Tamogashev[1], Nicola Branchini, Víctor Elvira, Jens Lagergren, and Nikolay Malkin. In International Conference on Learning Representations, 2026 [4]

List of papers by the author not included in the thesis.

**Paper E** **On the bias of variational resampling**, Axel Finke, Oskar Kviman, Nicola Branchini, and Víctor Elvira. In International Conference on Artificial Intelligence and Statistics, 2026

**Paper F** **LN's $t$-test: a principled approach to $t$-testing in single-cell RNA sequencing**, Oskar Kviman, Seong-Hwan Jun, Hosein Toosi, Pedro Ferreira, and Jens Lagergren. In bioRxiv, 2025

**Paper G** **Indirectly parameterized concrete autoencoders**, Alfred Nilsson, Klas Wijk, Sai Bharath Chandra Gutha, Erik Englesson, Alexandra Hotti, Carlo Saccardi, Oskar Kviman, Jens Lagergren,

Ricardo Vinuesa Motilva, and Hossein Azizpour. In International Conference on Machine Learning, 2024

**Paper H** **Variational resampling**, Oskar Kviman, Nicola Branchini, Víctor Elvira, and Jens Lagergren. In International Conference on Artificial Intelligence and Statistics, 2024

**Paper I** **Improved variational Bayesian phylogenetic inference using mixtures**, Ricky Molén[1], Oskar Kviman[1], and Jens Lagergren. In Transactions on Machine Learning Research, 2024

**Paper J** **Vaiphy: a variational inference based algorithm for phylogeny**, Hazal Koptagel[1], Oskar Kviman[1], Harald Melin, Negar Safinianaini, and Jens Lagergren. In Advances in Neural Information Processing Systems, 2022

[1]Equal contribution.

# Contents

# Introduction

# 1 Introduction

I develop this thesis around a broad class of inference problems: density estimation of multi-modal target distributions that naturally occur in computational biology. In my setting, the difficulty is rarely just dimensionality, but the multi-modal distributions I care about can be geometrically structured and only defined up to a normalizing constant. In the kappa, I make this challenge concrete by introducing a new unnormalized target density with deliberately difficult geometric properties (Section 3.2), and I return to it as a controlled test case when comparing inference strategies.

My methodological starting point is variational inference, i.e. optimization-based inference over a family approximating distributions. When analytic update equations are not available, I move to black-box variational inference (BBVI), which uses stochastic gradient estimators of the objective function, the evidence lower bound (ELBO), and heavily reduces the need for model-specific derivations (Chapter 4). From there, I focus on increasing expressiveness through ensembles of variational approximations, and I formalize this connection by defining the multiple importance sampling ELBO (MISELBO) for uniformly weighted ensembles (Section 4.2). The conclusion in this component of the thesis is that, when a single approximation is too restrictive, combining independently learned components can improve approximation quality—provided that the components are meaningfully diverse. The potential performance gain of using ensembles is formalized in Section 4.2.1, which I then use to clear a long-lived misconception about the possible posterior approximation accuracy improvement when using mixtures.

Ensemble learning can be fragile because their diversity and coverage depend on initialization and, heavily, on the maxima of the single-component objective. This motivates the next step: variational mixtures, which I explicitly distinguish from ensembles by whether the components are optimized jointly under a shared objective (Section 4.3). In the thesis, this trajectory is linked to two practical obstacles that prevent rich mixtures from being widely adopted: parameter growth in amortized settings and computational bottlenecks, including mixture-entropy costs that can scale poorly with the number of components. I therefore devote Section 4.4 to efficient variational mixture learning.

I then continue the methodological trajectory by moving from variational inference to flow matching, a framework for learning vector fields whose induced dynamics transform one probability distribution into another. After reviewing standard, bi-marginal flow matching, I focus on the necessity of specialized methods in multi-marginal problems and explain how the final contributed paper targets these challenges (Section 5.3). Finally, by combining methods and insights across the contributed papers, I derive a new multi-marginal flow matching method in Section 5.5: multi-marginal flow matching with mixtures of variational interpolants, and I introduce MVI-CFM, a component-aware loss designed to prevent the learned vector field from averaging out mixture information into meaningless trajectories.

## 1.1 Non-technical motivation

I like to explain the motivation for this thesis with a concrete example: imagine slicing a tumor into many thin layers, photographing each layer, and placing thousands of tiny "pins" into each photo, with each pin marking a location where you can read out molecular activity. That is close to what spatial transcriptomics provides: measurements of gene expression together with spatial location in a tissue section, often accompanied by stained microscopy images that help interpret the tissue and align sections. When I stack several sections from the same tissue, the slices become a kind of three-dimensional story: I can ask how spatial patterns evolve from one layer to the next.

Once I look at these slices, the 3D object I want to model is not a single number or a single curve, it is a probability distribution over 3D space, because each slice gives me a cloud of spatial points (Section 5.4.1). The utility of modeling this object with a statistical approach is that statistical models can be accurate in the presence of uncertainty. Here we are highly uncertain about the geometry of the full 3D object as we only obtain a small number of glimpses (the slices) of the tumor.

In Chapter 5, I rely on recent findings in the biological literature [9] which demonstrate that tumor volumes exhibit branching or looping geometries. This naturally gives rise to multi-modal distributions of tumor coordinates in space, and promotes the use of statistical models that are expressive enough to accurately predict the appearance of these multi-modal distributions in the unobserved, intermediate space. This is a major motivation of why the thesis cares about both multi-modality and how distributions dynamically evolve.

In *Multiple importance sampling ELBO and deep ensembles of variational approximations* (Paper A; [1]), I develop a way to combine several independently learned approximations (an ensemble) and evaluate them through the MISELBO objective. The practical motivation is that when there are multiple true answers, a single approximation can be too rigid, while a collection can cover more of what matters. In *Cooperation in the latent space: The benefits of adding mixture* *components in variational autoencoders* (Paper B; [2]), I study how mixture components can cooperate in order to explore different solutions that make sense given the data. This is exactly what I would want when the spatial organization of a tumor produces multiple distinct regions in a slice.

In *Efficient mixture learning in black-box variational inference* (Paper C; [3]), I address the question "yes, but can we afford it?" Mixture methods can become costly as the number of components grows, and my kappa explicitly discusses scaling concerns and introduces an amortization scheme intended to make mixture learning more efficient. Then, in *Multi-marginal flow matching with adversarially learnt interpolants* (Paper D; [4]), I turn to the second half of the story, namely learning how distributions dynamically evolve, i.e., connecting to the 3D spatial transcriptomics example, how distributions change across multiple slices.

Finally, the thesis culminates in a combined method in Section 5.5, where I combine all of the proposed methodologies and insights from Paper A-D to derive a new method *multi-marginal flow matching with mixtures of variational interpolants*. The new method is thus excellently posed to model multi-modal dynamics in general, and 3D spatial transcriptomics dynamics specifically, and naturally invites further investigation.

## 1.2 Thesis outline

To showcase the utility and flexibility of the methodologies to which my four papers contribute, and to fit a red thread that connects my works, I have decided to structure my thesis according to the following design choices.

**Drawing a methodological path** There is a trajectory that can be drawn from statistical inference development in the 1990's to today's generative artificial intelligence research. I will show where along this methodological path my works sit.

**Deriving a new method** At the end of the trajectory, in Section 5.5, I show how to naturally extend it by combining my four works [1, 2, 3, 4] into a new method. In other words, my thesis can be seen as a careful derivation of a new generative model, evolving side-by-side with the methodological path.

**A recurring inference task** I will propose a new, unnormalized target probability density with particular properties that makes inference challenging. The example will be revisited for each of the methodologies studied in Chapter 4.

# 2 Background

## 2.1 Notation

This chapter introduces the technical material required for the remainder of the thesis by first establishing notation and mathematical concepts that will be used throughout, including expectations, divergence measures, pushforwards, couplings, and vector fields.

The chapter also provides the biological context motivating several of the applications considered later, with a brief overview of spatial transcriptomics technologies and the types of probabilistic inference problems that arise in this setting. Together, these concepts form the conceptual foundation for the methodological developments presented in the subsequent chapters.

**Distributions and densities.** With abuse of notation, we use distributional symbols to denote probability density functions (pdfs) or probability mass functions (pmfs), depending on context. For example, if

$$p_\theta(x) = \frac{1}{\sqrt{2\pi\sigma^2}} \exp\left\{-\frac{(x-\mu)^2}{2\sigma^2}\right\} \tag{2.1.1}$$

is the pdf of a Gaussian random variable $x$ with parameters $\theta = \{\mu, \sigma^2\}$, then we write $p_\theta(x) = \mathcal{N}(x \mid \mu, \sigma^2)$. The parameter subscript may occasionally be dropped. When describing from which distribution a random variable is simulated, we write interchangeably $x \sim p_\theta(x)$ or $x \sim \mathcal{N}(x \mid \mu, \sigma^2)$. For ease of exposition in derivations, we may assume one-dimensional variables without loss of generality.

**Expectations and divergence measures.** We write expectations as $\mathbb{E}_p[f(X)]$ when $X \sim p$, and we drop the random-variable symbol when convenient, e.g. $\mathbb{E}_p[f(x)]$. The

(forward) Kullback-Leibler (KL) divergence between distributions $q$ and $p$ is

$$\mathrm{KL}(q\|p) = \mathbb{E}_q\left[\log \frac{q(z)}{p(z)}\right], \tag{2.1.2}$$

and we use $\mathbb{H}[q] = -\mathbb{E}_q[\log q]$ for the (differential) entropy. We use $\propto$ to denote equality up to a multiplicative constant that does not depend on the variable of interest. We also use the Radon-Nikodym derivative to reformulate integrals when convenient.

**Datasets and empirical measures.** We denote a dataset by $\mathcal{D} = \{x_i\}_{i=1}^n$ and $\delta_x$ as the Dirac measure at $x$. We informally treat $\mathcal{D}$ as a sample-based representation of an underlying distribution, and we switch between measure- and density-level notation when this does not cause ambiguity.

**Pushforwards and couplings.** If $x \sim p$ and $y = f(x)$ is a measurable map $f$, then the distribution of $y$ is denoted $f_{\#}p$ (the pushforward of $p$ through $f$). For joint distributions over pairs $(x, y)$, we write $\pi(x, y)$, and we use the term *coupling* when $\pi$ has prescribed marginals. In particular, if $x \sim p$ and $y \sim q$, then $\pi$ is a coupling between $p$ and $q$ if its marginals satisfy $\int \pi(\cdot, y)dy = p$ and $\int \pi(x, \cdot)dx = q$.

**Time-indexed collections.** For sequences of distributions we use indices $t_i$ (not necessarily representing physical time) and write

$$x_{t_i} \sim p_{t_i}, \qquad 0 = t_1 < t_2 < \cdots < t_K = 1. \tag{2.1.3}$$

We use the same notation for datasets $\mathcal{D}_{t_i}$ consisting of samples from $p_{t_i}$.

**Vector fields and flows.** When discussing deterministic dynamics, we use $v_t : \mathbb{R}^d \to \mathbb{R}^d$ for a time-dependent vector field and $\psi_t$ for the corresponding flow map solving the ODE $\frac{d}{dt}\psi_t(x) = v_t(\psi_t(x))$ with $\psi_0(x) = x$. The induced probability path from an initial distribution $p_0$ is denoted $p_t = (\psi_t)_{\#}p_0$.

**Asymptotic and linear-algebra notation.** We use $I_d$ for the $d \times d$ identity matrix, $\|\cdot\|$ for the Euclidean norm, $B^\top$ for a transposed matrix $B$ and use superscripts to index matrices, e.g. $\Sigma^{ij}$ denotes the $i$-th row and the $j$-th column in matrix $\Sigma$. The symbol $\mathcal{O}(\cdot)$ denotes unit-cost complexity.

## 2.2 Biological applications and data modalities

This thesis is motivated by inference problems that arise when samples are observed from complex biological systems. Specifically, we discuss inference of phylogenetic trees, and, more in-depth, consider spatial transcriptomics data.

### 2.2.1 Phylogenetic trees

A phylogeny is a tree that represents an evolutionary history. The leaves correspond to observed entities (e.g., species, samples, or cells), and internal nodes represent latent ancestors. Edges encode parent-child relationships, and branch lengths typically quantify evolutionary distance in time or in expected number of mutations.

Observations may be derived from DNA sequencing data, and the inferential goal is commonly to recover a posterior distribution over tree topologies and branch lengths.

This problem is challenging for approximate inference for several reasons. First, the space of tree topologies is combinatorial. Second, the posterior distributions are typically high-dimensional due to continuous branch lengths and latent ancestral states. Third, likelihood evaluations may be expensive and do not straightforwardly admit closed-form updates [10], which makes classical coordinate-ascent variational inference difficult to apply outside restricted conjugate settings. These properties motivate flexible black-box inference methodologies, which form the methodological starting point of this thesis.

### 2.2.2 Spatial transcriptomics

Spatial transcriptomics (ST; [11]) technologies measure gene expression while preserving spatial information in a tissue section. Operationally, a tissue slice is placed on an array of spatially indexed features (spots), and for each spot, a two-dimensional coordinate and a gene expression profile is returned. ST platforms also commonly provide a stained microscopy image (e.g. H&E), which can be used for visual interpretation and downstream alignment multiple sections from the same tissue.

In the context of this thesis, we primarily use the spatial coordinates to define distributions of interest. For a given tissue section (indexed by $i$), we may regard the set of observed coordinates as samples $x_{t_i} \sim \nu_{t_i}$ from a two-dimensional distribution over space. When several sections are obtained from a common tissue, the sections can be interpreted as a sequence (Equation (2.1.3)) along a third spatial axis, and one can ask how spatial patterns evolve across sections.

A critical practical consideration is that separate sections are not naturally aligned. Each slice is cut, stained, and imaged independently, which introduces rotations, scalings, and local distortions. As a consequence, the reported coordinates are typically expressed in different local coordinate systems, and an alignment step is needed before cross-section comparisons are meaningful [12]. Once aligned, the sequence of sections provides a set of sample-based spatial distributions that can be used to study interpolation and reconstruction tasks, such as predicting a held-out intermediate section or modeling a continuous three-dimensional spatial geometry from a few two-dimensional slices (see Section 5.4.1). Although our work in Kviman et al. [4] is unique in casting 3D ST as a prediction task, research on 3D ST is an active field, with examples including Mo et al. [9] and Liu, Zeira, and Raphael [13].

# 3 Variational Inference

This chapter introduces variational inference (VI; [14]), a framework for approximating intractable posterior distributions through optimization [15]. After presenting the evidence lower bound formulation and the classical coordinate ascent variational inference algorithm, we discuss the limitations that arise when analytic update equations are unavailable.

To study variational methods beyond analytically tractable settings, the chapter introduces a new unnormalized probability distribution constructed for this thesis. The distribution has geometric properties that make approximate inference challenging while remaining interpretable. It therefore serves as an example of when coordinate ascent variational inference is not applicable, and a controlled test for evaluating the flexible inference methodologies developed in the following chapter.

## 3.1 Mean-field variational approximations and the ELBO

In a 1995 NeurIPS paper, Lawrence K. Saul and Michael I. Jordan [16] obtained a lower bound on the intractable normalizing constant of a Boltzmann distribution—a textbook case of an intractable distribution in statistical inference, and one that historically motivated the development of the first Markov chain Monte Carlo (MCMC) method [17]. The lower bound was constructed with the goal of minimizing the Kullback-Leibler (KL) divergence from an approximating mean-field distribution,

$$q_\phi(z|x) = \prod_{i=1}^{m} q_i(z_i|x, \phi), \tag{3.1.1}$$

with $\phi$ the variational parameters and $m$ the number of dimensions in the latent variable, to a target posterior distribution $p_\theta(z|x)$ with parameters $\theta$, i.e.,

$$\mathrm{KL}(q_\phi(z|x) \| p_\theta(z|x)) = \mathbb{E}_{q_\phi(z|x)}\left[\log \frac{q_\phi(z|x)}{p_\theta(z|x)}\right]. \tag{3.1.2}$$

Since many posteriors of interest can be intractable due to their normalizing constants, much like the Boltzmann distribution, this KL divergence cannot always be computed directly. Instead, by writing the posterior via Bayes' rule

$$p_\theta(z|x) = \frac{p_\theta(z,x)}{p_\theta(x)} \tag{3.1.3}$$

and noting that the marginal likelihood $p_\theta(x)$, a.k.a. the evidence, is the normalizing constant, it was demonstrated that minimizing the KL divergence

$$\mathrm{KL}(q_\phi(z|x)\|p_\theta(z|x)) = -\mathbb{E}_{q_\phi(z|x)}\left[\log \frac{p_\theta(z,x)}{q_\phi(z|x)}\right] + \log p_\theta(x) \tag{3.1.4}$$

is equivalent to maximizing the tractable lower bound on the (log-) evidence

$$\begin{aligned}\log p_\theta(x) &= \mathbb{E}_{q_\phi(z|x)}\left[\log \frac{p_\theta(z,x)}{q_\phi(z|x)}\right] + \mathrm{KL}(q_\phi(z|x)\|p_\theta(z|x)) & (3.1.5)\\ &\geq \mathbb{E}_{q_\phi(z|x)}\left[\log \frac{p_\theta(z,x)}{q_\phi(z|x)}\right] = \mathcal{L}_{\mathrm{ELBO}}, & (3.1.6)\end{aligned}$$

hence the evidence lower bound (ELBO).

Although the construction of this lower bound and the mean-field approach were greatly inspired by the statistical physics research field, [16] is the first machine-learning-centric work where these concepts appear, and it clearly marks the start of variational inference (VI) research.

**Coordinate ascent variational inference** The mentioned work, along with those that followed [18, 19, 20], were focused on inference problems in probabilistic graphical models, such as Bayesian nets, and the ELBO maximization algorithms were developed accordingly. However, by later connecting VI to the expectation-maximization algorithm [21], ELBO maximization via coordinate ascent was developed in Jordan et al. [14], effectively giving rise to coordinate ascent VI (CAVI).

Under a mean-field factorization (see Equation (3.1.1)), CAVI maximizes the ELBO by iteratively optimizing each factor $q_i$ while holding the remaining factors fixed. Taking derivatives of the ELBO w.r.t. $q_i$ yields the update equation

$$\log q_i^\star(z_i|x,\phi) \propto \mathbb{E}_{q_{-i}(z_{-i}|x,\phi)}\left[\log p_\theta(z_i, z_{-i}, x)\right], \tag{3.1.7}$$

where $q_{-i} = \prod_{j\neq i} q_j$ and the constant ensures normalization. Thus, each factor is updated to match $p_\theta(z_i, z_{-i}, x)$, marginalized over the remaining variables w.r.t. the current variational distribution.

When the conditional distributions, $p_\theta(z_i|z_{-i}, x)$, and each factor of the variational distribution, $q_i(z_i|x,\phi)$, belong to the same tractable exponential family, these updates enable fast inference via *conjugacy*, yielding closed-form variational parameter updates [22, 23].

**Intractability of CAVI** Unfortunately, Equation (3.1.7) is intractable or requires tedious derivations for many families of target distributions, but also variational families. In a technical note by Neil D. Lawrence from 2002 [24], he proposed an idea in this setting by estimating Equation (3.1.7) using an importance sampler.

Unaware of the existence of the technical note until later,[1] we developed variational importance samplers in our CAVI algorithm for phylogenetic tree inference [10] to estimate intractable update equations. In the extension of [10] to tumor phylogenies [25], the variational importance sampler was also used.

The variational importance sampler is of course only one approximate inference solution to the problem of the intractable update equations in Equation (3.1.7), however, every approximation scheme introduces new challenges that reduce the practical appeal of CAVI when requiring complex distribution families.

## 3.2 A new probability distribution

To study variational methods beyond conjugate and analytically tractable settings, I introduce an unnormalized density constructed for this thesis that does not allow for CAVI updates and whose geometry is sufficiently rich to make approximate inference challenging. The resulting target distribution necessitates flexible variational approximations, for which closed-form CAVI updates do not exist. At the same time, it serves as a controlled experiment for benchmarking the methods presented in the next chapter.

The proposed distribution models correlated random variables on a unit hypercube, and is inspired by an observation we made in Kviman et al. [6]. Namely, there it was shown that a uni-variate log-normal (LN) pdf

$$\mathrm{LN}(\theta|\mu,\sigma^2) = \frac{1}{\theta\sqrt{2\pi\sigma^2}} \exp\left\{\frac{-(\log\theta-\mu)^2}{2\sigma^2}\right\} \tag{3.2.8}$$

can be parameterized to closely mimic the Beta pdf

$$\mathrm{Beta}(\theta|a,b) = \frac{\theta^{a-1}(1-\theta)^{b-1}}{\mathrm{B}(a,b)}, \tag{3.2.9}$$

with B the Beta function, when $a \leq b$ (i.e., when most mass is distributed on the region $[0, 0.5]$). However, regardless of $a$ and $b$ are chosen, the LN pdf inevitably distributes mass outside the unit line, which might be an unacceptable property for certain modeling tasks (e.g. when $\theta$ corresponds to the probability of an event).

Here we extend the LN approximation to a new unnormalized pdf with support on a $d$-dimensional hypercube $(0,\infty)^d \cap (-\infty,1)^d = (0,1)^d$, which allows for correlation modeling via LN covariance matrices, $\Sigma \in \mathbb{R}^{d\times d}$. The intuition behind the construction of the distribution is that the support of the product of two *colliding* LN distributions will be upper and lower bounded by the LN truncations, inspiring the following expression for the colliding LN (CoLN; *Collin*) distribution

$$\text{CoLN}(\theta|\mu_1, \mu_2, \Sigma_1, \Sigma_2) = \text{LN}(\theta|\mu_1, \Sigma_1)\text{LN}(1-\theta|\mu_2, \Sigma_2), \tag{3.2.10}$$

which has support on $(0, 1)^d$ as $\text{LN}(\theta|\mu_1, \Sigma_1)$ and $\text{LN}(1-\theta|\mu_2, \Sigma_2)$ are defined on $(0, \infty)^d$ and $(-\infty, 1)^d$, respectively.

[1] The existence of the *variational importance sampler* was made aware to me by Matthew D. Hoffman who pointed me to the mentioned technical note during a chat at the AISTATS conference.

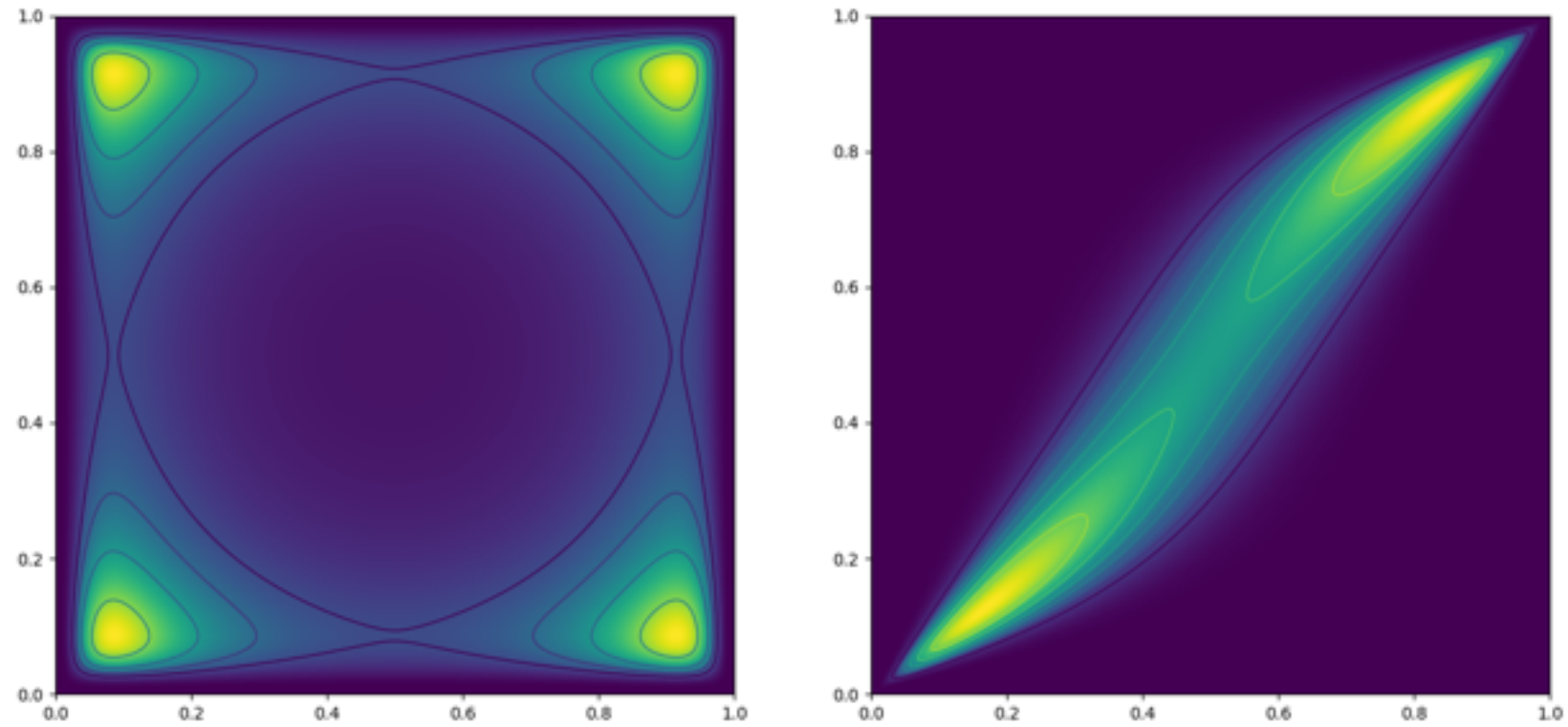

**Figure 3.2.1:** Contour plots of two different parameterizations of the CoLN distribution. The quad-modal distribution (left) is parameterized with $\exp(\mu) = [0.1, 0.1]^\top$ and $\Sigma = 0.4I$, while the bi-modal distribution uses $\exp(\mu) = [0.5, 0.35]^\top$, $\Sigma^{11} = \Sigma_{22} = 0.65$ and $\Sigma^{12} = \Sigma_{21} = 0.6$.

The CoLN distribution has an intractable normalizing constant due to the product of two non-linear transformations of $\theta$, which prevents sampling from it directly. But if we assume that $\mu = \mu_1 = \mu_2$ and $\Sigma = \Sigma_1 = \Sigma_2$ the expression for the unnormalized pdf simplifies to

$$\text{CoLN}(\theta|\mu, \Sigma) = \frac{e^{-\frac{1}{2}\left[(\log\theta-\mu)^\top\Sigma^{-1}(\log\theta-\mu)+(\log(1-\theta)-\mu)^\top\Sigma^{-1}(\log(1-\theta)-\mu)\right]}}{(2\pi)^d|\Sigma|\prod_{i=1}^d \theta_i(1-\theta_i)}, \tag{3.2.11}$$

and the geometry of the distribution is sufficiently interpretable to construct two interesting, multi-modal distributions (in $d = 2$ for the sake of visualization): for a diagonal covariance matrix with small diagonal elements ($\Sigma = 0.4I$) and small means in log space ($\mu = \log([0.1, 0.1]^\top)$), there are four modes approximately in $\{0.1, 0.9\} \times \{0.1, 0.9\}$, shown in the left plot in Figure 3.2.1. Intuitively, the resulting distribution has a similar geometry to two identical Beta marginal distributions with $a, b < 1$ (see Equation (3.2.9)).

Beta distributions do not allow for correlation modeling between marginal distributions, or, equivalently, across dimensions on the hypercube. However, we can simply choose non-zero off-diagonal entries in the covariance matrix to achieve this. For instance (right plot in Figure 3.2.1), letting $\mu = \log([0.5, 0.35]^\top)$, $\Sigma^{11} = \Sigma^{22} = 0.65$ and $\Sigma^{12} = \Sigma^{21} = 0.6$ the CoLN distribution is bi-modal with a correlation structure on $(0, 1)^2$.

The goal of introducing the CoLN distribution is not to exhaustively analyze its theoretical properties, but to construct an unnormalized density whose geometry is sufficiently rich to make approximate inference genuinely challenging. The multi-modality, bounded support and covariance structure of the two distributions in Figure 3.2.1, are precisely examples of features that render the CAVI update equations intractable and motivate the methods presented in the upcoming chapter. We conclude the chapter by situating the proposed construction within the existing literature.

**Related work** The construction principle of multiplying densities is well-known and was popularized in machine learning by Geoffrey E. Hinton via the product of experts framework [26], but had been considered earlier in Bayesian statistics under the name logarithmic opinion pooling [27]. These are, on the other hand, general density construction frameworks, and we could not find the CoLN distribution as an established instance of this class.

Based on its finite support on the unit hypercube, it is natural to investigate connections to distributions or methods that (i) generalize the Beta distribution to a multi-variate counterpart, and (ii) have support on the unit hypercube.

The arguably most famous example of methods that formalize the construction of correlated marginal distributions is copulas [28, 29]. Compactly explained, one samples a multi-variate Gaussian random variable with covariance structure, and applies a dimension-wise transformation using a uni-variate Gaussian cumulative density function (cdf) to obtain uniform draws. Finally, apply marginal Beta inverse cdfs to get a multi-variate random variable with correlation structure according to the multi-variate Gaussian (base) distribution. Clearly, the sample generated via the copula (the cdf transformation scheme) has Beta distributed marginals, which is one of multiple distinctions to CoLN which can have non-Beta marginals.

Two benefits with copulas over CoLN are that the resulting distributions are normalized and that it is possible to sample from them. Unfortunately, however, the Beta inverse cdf is notoriously not available in closed form, making it inconvenient in practice. The Kumaraswamy distribution [30] enjoys a closed form inverse cdf and has often been promoted as an alternative to the Beta distribution when required to work with the Beta inverse cdf [31]. For future work, if one parameterizes the CoLN distribution with the purpose to approximate the Beta marginal pdfs, then one could study the relation between CoLN, a Kumaraswamy copula and a Beta copula.

In the second category of related works, i.e. distributions on the unit hypercube, we find a range of distributions extended based on the generalized Beta distribution [32], however, these models are proposed for the unit square [33] (that is, bi-variate Beta distributions) although multi-variate extensions are discussed [34]. Indeed, as indicated by the bi-variate Beta design guide in Olkin and Trikalinos [35] and through some of the recent publication dates, multi-variate Beta construction is an established and active research field. Nonetheless, CoLN does not fall into the category of generalized Beta distributions.

Finally, the logit-Gaussian [36, 37], where a logit-transformed random variable is assumed Gaussian, is another approach to obtain multi-variate distributions with correlation structure on a unit hypercube [38]. Without need for further justification, a logit-Gaussian random variable follows a distribution fundamentally different from the CoLN distribution.

# 4 Black-Box Variational Inference

This chapter moves from classical variational inference to black-box variational inference (BBVI; [39]), a class of methods designed to handle models and variational families for which analytic update equations are unavailable, and for reducing tedious derivation steps in general. Instead of relying on model-specific derivations, BBVI uses stochastic gradient estimators of the ELBO (Equation (3.1.6)) that can easily be applied to a wide range of models.

We review the key ideas behind BBVI and then focus on methods that increase the expressiveness of variational approximations, in particular ensembles of variational approximations [1] and variational mixtures [2, 3].

## 4.1 Black-box variational inference

In a talk I attended by Rajesh Ranganath when the paper *Black Box Variational Inference* [39] by him, Sean Gerrish and David Blei received the test-of-time award at AISTATS-24, Rajesh described that the motivation for developing a black-box alternative to CAVI was the latter's inflexibility: simply changing the target distribution will in many cases result in great overhead in terms of calculations and derivations, and the CAVI update equations might even be intractable for many families of target or variational distributions.

Instead, BBVI was introduced as a flexible methodology which does not require explicit update equations by instead Monte-Carlo estimating the gradient of the ELBO objective (Equation (3.1.6)) via an unbiased gradient estimator

$$\nabla_\phi \mathbb{E}_{q_\phi(z|x)}\left[\log \frac{p(x,z)}{q_\phi(z|x)}\right] = \mathbb{E}_{q_\phi(z|x)}\left[\log \frac{p(x,z)}{q_\phi(z|x)} \nabla_\phi \log q_\phi(z|x)\right] \tag{4.1.1}$$

$$\approx \frac{1}{M}\sum_{m=1}^{M} \log \frac{p(x,z^{(m)})}{q_\phi(z^{(m)}|x)} \nabla_\phi \log q_\phi(z^{(m)}|x), \tag{4.1.2}$$

where the $z^{(m)}$ samples are i.i.d. draws from $q_\phi(z|x)$. This framework is agnostic to the choice of $p(x, z)$ and $q_\phi(z|x)$ as long as the (potentially unnormalized) target and the variational distribution can both be evaluated, we can sample from the variational distribution, and, of course, if the ELBO is defined (e.g., $p(z, x) > 0$ for all $z \sim q_\phi(z|x)$).

Concurrently as BBVI [39] was published at AISTATS, Kingma and Welling proposed the variational autoencoder (VAE; [40]) at ICLR—the two conference submission deadlines are typically separated by merely a couple of weeks. The two works are in many regards solving the same flexibility issues with CAVI, but bring some distinct contributions. The work in Ranganath, Gerrish, and Blei [39] displayed the general applicability of BBVI to the extent that the VAE can be seen as an instance of the BBVI methodology, and inference was not constrained to continuous or discrete random variable settings. Meanwhile, in Kingma and Welling [40], the reparameterization trick and amortized inference of distributional parameters were introduced, two concepts that underpin the developmental success of generative AI.

**The reparameterization trick** A challenge with applying gradient-based inference of the variational parameters is that the ELBO is an expectation w.r.t. the variational distribution. In Equation (4.1.1), the issue is solved via the construction of an unbiased gradient estimator which is applicable for both discrete and continuous latents ($z$), but, unfortunately, the estimator suffers from high variance [39, 41, 42]. A low-variance alternative proposed in Kingma and Welling [40] was the reparameterization trick. Assuming a Gaussian variational distribution, $q_\phi(z|x) = \mathcal{N}(\mu, \sigma^2)$, i.e. $\phi = (\mu, \sigma^2)$, then reparameterization of the Gaussian,

$$z_\phi(\epsilon) = \mu + \sigma\epsilon, \quad \epsilon \sim p(\epsilon) = \mathcal{N}(0, 1) \tag{4.1.3}$$

results in an expectation w.r.t. a distribution that does not depend on $\phi$ and so $\nabla_\phi$ can be pushed inside the expectation:

$$\nabla_\phi \mathbb{E}_{q_\phi(z|x)}\left[\log \frac{p(x,z)}{q_\phi(z|x)}\right] = \mathbb{E}_{p(\epsilon)}\left[\nabla_\phi \log \frac{p(x, z_\phi(\epsilon))}{q_\phi(z_\phi(\epsilon)|x)}\right]. \tag{4.1.4}$$

The trick was later extended to categorical distributions [42, 43] and has been widely adopted in BBVI [44, 3], normalizing flows [45] (introduced below), as well as state-of-the-art generative AI methodologies like Gaussian probability path flow matching [46] (introduced in the subsequent chapter).

**Amortized inference** When Matthew D. Hoffman spoke as an invited speaker at AISTATS 2024, he compared different aspects of VI vs. MCMC methods. As I recall the talk, amortized inference was described as the major favorable feature of the VI methodology.

The standard MCMC methods[1] and non-amortized VI methods (like CAVI) sample from or approximate posterior distributions based on a single dataset. Meanwhile, in amortized inference a general mapping from data to parameters is learned based on a dataset. Since the mapping is shared across all samples in the dataset, it can map new data points to the parameter space—without additional inference. In a VAE this mapping is an encoder network, i.e., the amortized inference map is a neural network.

[1]The MCMC literature is vast, and so there are likely methods where amortization is applied. However, the general and widely applied MCMC methods do not straightforwardly admit amortization.

Amortized inference maps allow for generalization to unobserved data, and their generalization properties are crucial for sampling in generative AI models—in flow matching the vector field is learned via an amortization schemes w.r.t. time, which allows for down-stream sampling via simulation of the ordinary differential equation with arbitrary step sizes.

**Normalizing flows** Although BBVI made it possible to be more flexible in selecting both the target and the variational distribution families, the ease of applying the reparameterization trick naturally promoted Gaussian variational approximations. This choice is of course an oversimplified approximation that will gravely mismatch any interesting posterior target distribution.

By exploiting the convenient properties of Gaussian distributions, [45] elegantly employed the change-of-variables theorem to construct a series of $K$ transformations, $T_{1:K}$, that maps a simple Gaussian base distribution, $q_0(z)$, into an arbitrarily complex distribution with log-density

$$\log q_K(z_K) = \log q_0(z_0) - \sum_{i=1}^{K} \log \left| \frac{\partial T_i}{\partial z_{i-1}} \right|. \tag{4.1.5}$$

The transformations are chosen as bijective and invertible functions which were trained in Rezende and Mohamed [45] by ELBO maximization, relying on the reparameterization trick. This sequence of transformations is known as a normalizing flow (NF). While certain classes of NFs are universal density approximators, their expressivity can depend critically on depth, and shallow flows may fail to approximate some target distributions [47, 48].

## 4.2 Ensembles of variational approximations

A straightforward alternative approach for avoiding the oversimplified Gaussian variational family is to construct an ensemble of independently learned variational distributions (Paper A; [1]). That is, we can obtain more sophisticated approximations by inferring the variational parameters of $A$ components that independently maximize the ELBO, and then construct an ensemble with these components,

$$q_{\phi_{1:A}}(z|x) = \frac{1}{A} \sum_{a=1}^{A} q_{\phi_a}(z|x). \tag{4.2.6}$$

This is indeed traditionally called a mixture model, but we make a distinction between variational ensembles and variational mixtures based on their different inference schemes.

**Definition 1.** *Both mixtures and ensembles are weighted combinations of components. If the component parameters are optimized separately, the resulting distribution is an ensemble of variational approximations. If they are optimized jointly under a common objective, it is a variational mixture.*

Variational mixtures are the topic of the next section.

Motivated by the connection between ensembles of variational distributions and multiple-importance samplers [49], we termed a definition of the ELBO for uniformly weighted ensembles as the multiple importance ELBO (MISELBO),

$$\mathcal{L}_{\text{MIS}}(q_{\phi_{1:A}}(z|x)) = \frac{1}{A}\sum_{a=1}^{A}\mathbb{E}_{z_a\sim q_{\phi_a}(z|x)}\left[\log\frac{p_\theta(x,z_a)}{\frac{1}{A}\sum_{a'=1}^{A}q_{\phi_{a'}}(z_a|x)}\right], \tag{4.2.7}$$

which naturally admits the importance weighted ELBO extension [50], namely

$$\mathcal{L}^{L}_{\text{MIS}}(q_{\phi_{1:A}}(z|x)) = \frac{1}{A}\sum_{a=1}^{A}\mathbb{E}_{z_{a,\ell}\sim q_{\phi_a}(z|x)}\left[\log\frac{1}{L}\sum_{\ell=1}^{L}\frac{p_\theta(x,z_{a,\ell})}{\frac{1}{A}\sum_{a'=1}^{A}q_{\phi_{a'}}(z_{a,\ell}|x)}\right], \tag{4.2.8}$$

where $\lim_{L\to\infty}\mathcal{L}^{L}_{\text{MIS}} = \log p_\theta(x)$. The importance weighted ELBO in Equation (4.2.8) is an important quantity in order to get marginal log-likelihood estimates, often $L$ in chosen in the order of thousands [50, 51].

### 4.2.1 Performance gain of ensembles

The density estimation accuracy gain of using ensembles over a single distribution—in terms of tighter ELBOs, and so smaller KL divergences—was theoretically determined in Kviman et al. [1] via the following theorem.
**Theorem 1.** *Define the average ELBO score of the ensemble components as*

$$\bar{\mathcal{L}}_{\text{ELBO}}(q_{\phi_{1:A}}) = \frac{1}{A}\sum_{a=1}^{A}\mathcal{L}_{\text{ELBO}}(q_{\phi_a}),$$

*where* $\mathcal{L}_{\text{ELBO}}(q_{\phi_a})$ *is given in Equation* (3.1.6)*, and let*

$$\Delta = \mathcal{L}_{\text{MIS}}(q_{\phi_{1:A}}) - \bar{\mathcal{L}}_{\text{ELBO}}(q_{\phi_{1:A}}),$$

*then*

$$\Delta \in [0, \log(A)]. \tag{4.2.9}$$

The proof of the theorem given in Kviman et al. [1] is developed in two steps. First, establish that the gap between MISELBO and the average of separate ELBOs ($\Delta$) is exactly the Jensen-Shannon divergence (JSD)

$$\Delta = \mathcal{L}_{\text{MIS}}(q_{\phi_{1:A}}) - \frac{1}{A}\sum_{a=1}^{A}\mathcal{L}_{\text{ELBO}}(q_{\phi_a}) \tag{4.2.10}$$

$$= \mathbb{H}\left[q_{\phi_{1:A}}(z|x)\right] - \frac{1}{A}\sum_{a=1}^{A}\mathbb{H}\left[q_{\phi_a}(z|x)\right] = \text{JSD}\left(q_{\phi_{1:A}}(z|x)\right), \tag{4.2.11}$$

where $\mathbb{H}(\cdot)$ is the entropy operator. Then it suffices to note that $\text{JSD}(q_{\phi_{1:A}}(z|x)) \in [0, \log(A)]$.

From the characterization of $\Delta$ as the JSD, we get an upper bound ($\log(A)$) on the density estimation capacity improvement of ensembles over single variational distributions. This upper bound is achieved, for instance, when the ensemble components are sufficiently separated in the latent space such that $\sum_{a\neq a'} q_{\phi_{a'}}(z_a) = 0$ for all $z_a \sim q_a(z)$, which is easy to show:

**Theorem 2.** *Assume that* $\sum_{a\neq a'} q_{\phi_{a'}}(z_a) = 0$ *for all* $z_a \sim q_a(z)$*, then* $\Delta = \log(A)$.

*Proof.*

$$\mathcal{L}_{\text{MIS}}(q_{\phi_{1:A}}) = \frac{1}{A}\sum_{a=1}^{A} \mathbb{E}_{q_{\phi_a}(z)}\left[\log \frac{p_\theta(x,z)}{\frac{1}{A}\sum_{a'=1}^{A} q_{\phi_{a'}}(z)}\right] \tag{4.2.12}$$

$$= \frac{1}{A}\sum_{a=1}^{A} \mathbb{E}_{q_{\phi_a}(z)}\left[\log \frac{p_\theta(x,z)}{\frac{1}{A} q_{\phi_a}(z)}\right] \tag{4.2.13}$$

$$= \frac{1}{A}\sum_{a=1}^{A} \mathcal{L}(q_{\phi_a}) + \log(A), \tag{4.2.14}$$

and so $\Delta = \mathcal{L}_{\text{MIS}}(q_{\phi_{1:A}}) - \frac{1}{A}\sum_{a=1}^{A} \mathcal{L}(q_{\phi_a}) = \log(A)$. □

As implied by the assumption in the theorem above, larger diversity among the ensemble components leads to larger performance gains in terms of MISELBO scores. Given that the posterior geometry is sufficiently complex such that there are multiple, separated high-probability regions, the diversity among the components mainly stems from the initialization of the variational parameters, or stochastic optimization.

### 4.2.2 Related work

Given the simplicity and the provable performance gain of ensembles, there were surprisingly few examples of ensembles in VI when they were proposed in Kviman et al. [1]. They had only been explored to a very limited extent. From a black-box VI perspective, the most notable example of ensembles of variational approximations were considered in Lopez et al. [52] to form multiple importance samplers for various decision-making tasks, but the density estimation accuracies of the ensembles were not quantified as the ensemble was not the primary topic of the study.

Interestingly, when researching the early history of VI for this thesis, I found an informal statement relating to Theorem 1, made about the improvement of mixture models versus single components in mean-field VI [20]. There, the equivalence to $\Delta$ was instead expressed in terms of the mutual information between the ensemble component index $a$ and the latent variable $z$. This mutual information, as it is expressed in Jaakkola and Jordan [20], happens to be equivalent to the JSD

$$I(a;z) = \frac{1}{A}\sum_{a=1}^{A} \mathrm{KL}\left(q_{\phi_a}(z|x) \| q_{\phi_{1:A}}(z|x)\right) \tag{4.2.15}$$

$$= \frac{1}{A}\sum_{a=1}^{A} \mathbb{E}_{q_{\phi_a}(z|x)}\left[\log q_{\phi_a}(z|x)\right] - \frac{1}{A}\sum_{a=1}^{A} \mathbb{E}_{q_{\phi_a}(z|x)}\left[\log q_{\phi_{1:A}}(z|x)\right] \tag{4.2.16}$$

$$= -\frac{1}{A}\sum_{a=1}^{A} \mathbb{H}[q_{\phi_a}(z|x)] + \mathbb{H}[q_{\phi_{1:A}}(z|x)] = \mathrm{JSD}\left(q_{\phi_{1:A}}(z|x)\right), \tag{4.2.17}$$

and is therefore identical to $\Delta$ in Theorem 1. As mutual information satisfies $I(a;z) \leq \log(A)$, this yields the same upper bound as in our Theorem 1, that is, on the improvement relative to the average ELBO of the same component distributions. Regarding the result, Jaakkola and Jordan [20] write:

> *"It is the second term, i.e., the mutual information, that characterizes the gain of using the mixture approximation. As a non-negative quantity, $I(m;S)$ increases the likelihood bound and therefore improves the approximation. We note, however, that since $I(m;S) \leq \log M$, where $M$ is the number of mixture components, the KL divergence between the mixture approximation and the true posterior can decrease at most logarithmically in the number of mixture components."*

The last sentence above is only correct when interpreted as referring to the improvement of a mixture relative to the weighted average of ELBOs based on the same components from which the mixture is constructed. This is exactly what is formalized in our Theorem 1. However, the conclusion in the quote's last sentence above suggests a stronger and more general limitation on mixture approximations than what is formally established.

The scope of the Jaakkola and Jordan [20] result has occasionally been interpreted more broadly in the subsequent literature. For instance, [53] write that "*the KL divergence between the mixture distribution $q$ and the true posterior decreases at best logarithmically in the number of mixture components $N$,*" citing Jaakkola and Jordan [20]. This suggests that the approximation error of mixture variational families can improve at most logarithmically with the number of components.

However, it is important to interpret the original result correctly. The $\log(A)$ bound applies to the performance gain of a mixture relative to the *average of its constituent component ELBOs*. More precisely, the mutual-information term $I(a;z)$ quantifies the increment obtained by combining a fixed collection of component approximations into a mixture, and satisfies $0 \leq I(a;z) \leq \log(A)$. The bound therefore controls how much tighter the mixture ELBO can be compared to weighted average of the ELBO scores from the same components.

It does *not* imply that a mixture cannot outperform an arbitrary single variational approximation by more than $\log(A)$, nor does it impose a universal logarithmic ceiling on the KL divergence achievable by increasingly expressive mixture families. The mutual-information identity concerns the gain from mixing a given set of components, it does not characterize the global approximation capacity of variational mixture families.

We end this discussion by demonstrating by an illustrative counterexample below, showing that a mixture can indeed outperform certain single-component approximations by more than a logarithmic factor in KL.

**Counterexample to a logarithmic KL upper bound** Let

$$p_\theta(x,z) = \frac{\epsilon}{A}\mathcal{N}(z|\mu_1,\sigma_1^2) + \frac{A-\epsilon}{A}\mathcal{N}(z|\mu_2,\sigma_2^2), \quad \epsilon \in (0,1), \tag{4.2.18}$$

and assume that we perfectly fitted a weighted variational mixture to this target in the sense that

$$\mathcal{L}_{\text{MIS}} = -\text{KL}\left(q_{\phi_{1:A}}(z|x) \| p_\theta(x,z)\right) = 0. \tag{4.2.19}$$

Then assume a single Gaussian approximation that instead perfectly fits the lower-probability Gaussian, $q_{\phi'}(z|x) = \mathcal{N}(z|\mu_1,\sigma_1^2)$, and assume that the two modes in the target distribution are sufficiently separated, such that $\mathcal{N}(z|\mu_2,\sigma_2^2) \approx 0$ for all $z \sim q_{\phi'}(z|x)$, then

$$\mathcal{L}_{\text{ELBO}}(q_{\phi'}(z|x)) = \mathbb{E}_{q_{\phi'}(z|x)}\left[\log \frac{\frac{\epsilon}{A}\mathcal{N}(z|\mu_1,\sigma_1^2) + \frac{A-\epsilon}{A}\mathcal{N}(z|\mu_2,\sigma_2^2)}{q_{\phi'}(z|x)}\right] \tag{4.2.20}$$

$$\approx \mathbb{E}_{q_{\phi'}(z|x)}\left[\log \frac{\frac{\epsilon}{A}\mathcal{N}(z|\mu_1,\sigma_1^2)}{q_{\phi'}(z|x)}\right] = \log(\epsilon/A), \tag{4.2.21}$$

and so

$$\Delta = 0 - \log(\epsilon/A) = -\log(\epsilon) + \log(A) > \log(A) \quad \forall \epsilon \in (0,1). \tag{4.2.22}$$

This does not contradict the $\log(A)$ bound in Theorem 1, because that bound concerns the improvement of a mixture relative to the bounds of its own constituent components, not relative to an arbitrary single approximation chosen independently of the mixture construction.

### 4.2.3 Experiments

Here we analyze properties of ensembles of variational approximations by considering the two CoLN distributions in Figure 3.2.1 as target densities in a density estimation task. The two targets are quad- and bi-modal, respectively, and are both bounded on the unit rectangle. The bounded support causes truncation of the quad-modal distribution, while the bi-modal distribution has a non-linear covariance structure.

To estimate the targets, we use $A = 10$ components and let $q_{\phi_a}(z|x) = \mathcal{N}(z|\mu_a, \Sigma_a)$, with $\Sigma_a^{ij} = 0$ for all $a \in [1, A]$ and $i \neq j$. The ensemble weights are uniform. All

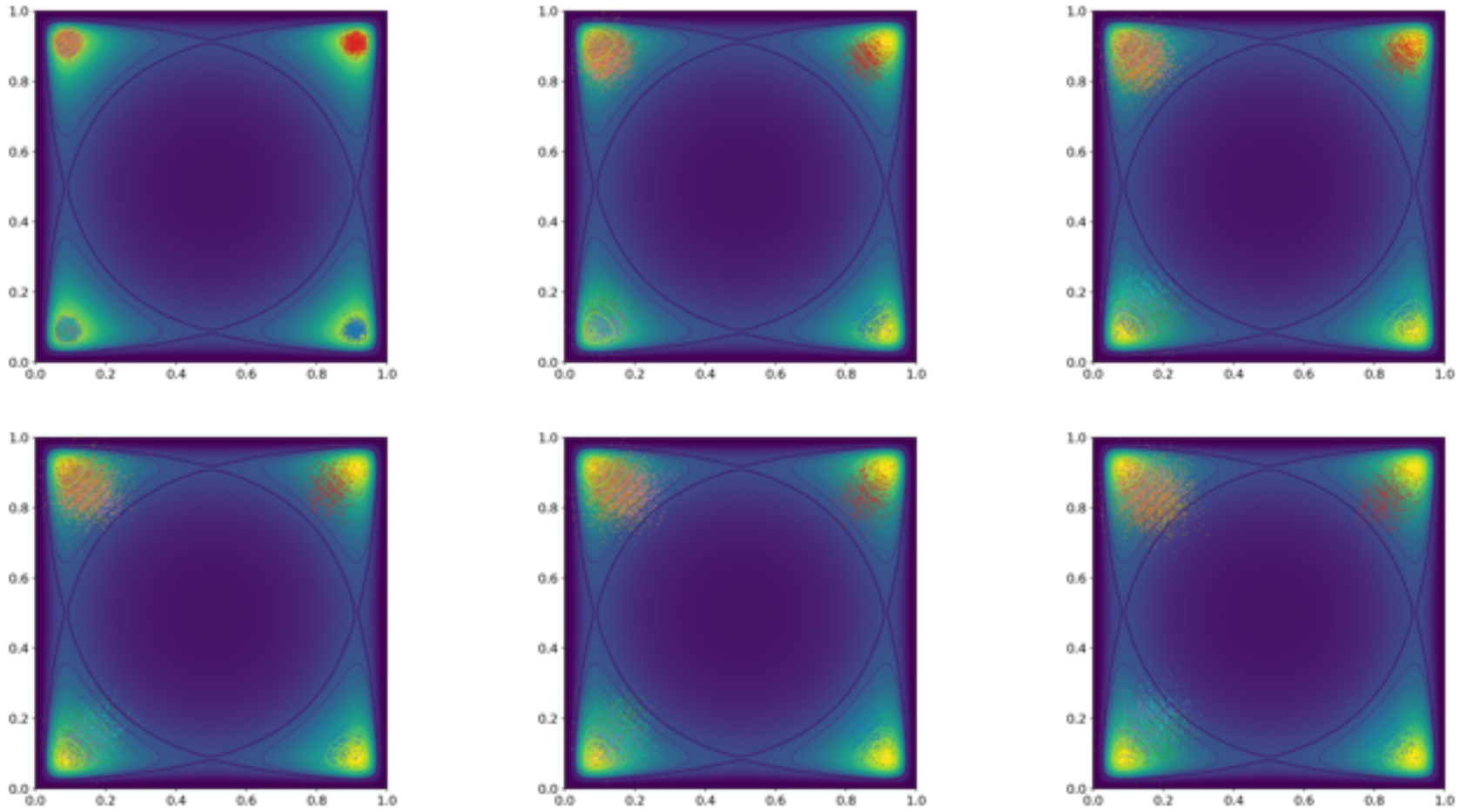

**Figure 4.2.1:** The evolution of the $A = 10$ single-component approximations after $1$k, $2$k, $3$k, $4$k, $5$k and $10$k training iterations, respectively. The component mean parameters are randomly initialized around $(0.5, 0.5)$ and individually maximize the ELBO objective to match the CoLN target distribution (its unnormalized density is shown via the contour plot in all figures; see Section 3.2 for details regarding CoLN). The colored scatter points are samples from the different components with separate colors for each component.

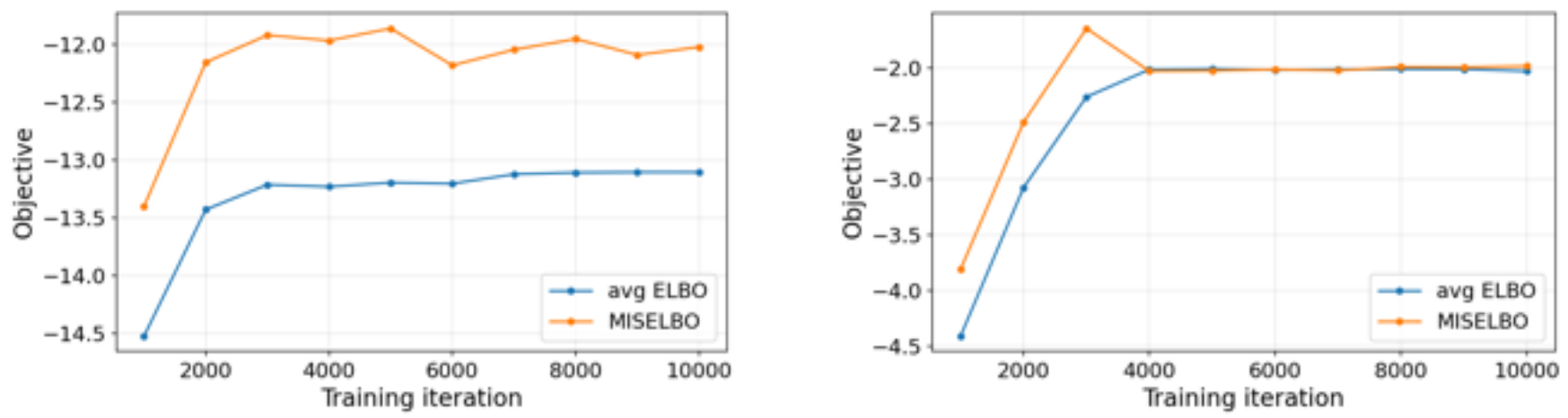

**Figure 4.2.2:** MISELBO and average (avg) ELBOs curves for the ensemble approximations on the quad-modal (left) and bi-modal (right) target densities.

component means are initialized randomly around $(0.5, 0.5)$, while the component variances are initialized to small values to avoid sampling outside the unit-rectangle.

For the quad-modal target density, the ensemble of variational approximation results are show in Figure 4.2.1 and the corresponding MISELBO vs. average single-component ELBOs curves are provided in Figure 4.2.2. The ten approximations randomly and independently start covering one of the four modes (see the left-most, top subfigure in Figure 4.2.1). The ELBO objective appears to incentivize increased variances of the individual components, but as the variances grow, the truncation of the target density forces the component means to shift away from the modes.

In Figure 4.2.3 we share the results for the bi-modal target case. The two modes are quickly found, and the component means are centered in the modes already after $1$k training iterations (left-most, top subfigure). However, again, the single-component ELBOs promote the component variances to grow, and after $2$k iterations (center, top subfigure), all component means are drawn back towards $(0.5, 0.5)$, until after $4$k training iterations, all components find the same optimum (bottom row). That the approximations in the final subfigure all represent the same optimal solution is verified by inspecting Figure 4.2.2, where we see that MISELBO and the average ELBOs ultimately plateau and coincide—all approximations are optimal and they have approximately the same parameters.

In both settings, the Adam optimizer was used [54] with default parameters, except a learning rate of $9 \times 10^{-4}$, to maximize the ELBOs.

To summarize the results in both experimental setups, we emphasize the simplicity of implementing ensembles of variational approximations and note that the result of using ensembles instead of single components is positive in the quad-modal experiment (observe the gap between the MISELBO curve and average ELBO curve).

On the other hand, the inflexibility of the diagonal Gaussian density forces the individually inferred approximations to sacrifice coverage of high-probability regions (modes), and the resulting approximations are clearly inaccurate. From Theorem 2 we know that the performance gain of using ensembles is maximized when the components are diverse, but apparently the single-component ELBOs in the considered problems do not result in diverse solutions.

In the next section we will show how to circumvent the inflexibility issues by achieving more diverse component sets, by introducing variational mixtures instead (see Definition 1).

## 4.3 Variational mixtures

The diversity of the ensemble components, and so the approximation capacity of the ensemble, can highly depend on initialization of the variational parameters. When the geometry of the target distribution is straightforward to analyze, for instance visually as in the Figure 3.2.1, inducing diversity via appropriate initialization is quite easy. However, if the target distribution is high-dimensional or the variational parameters are outputs from a neural network, then initialization-induced diversity is challenging.

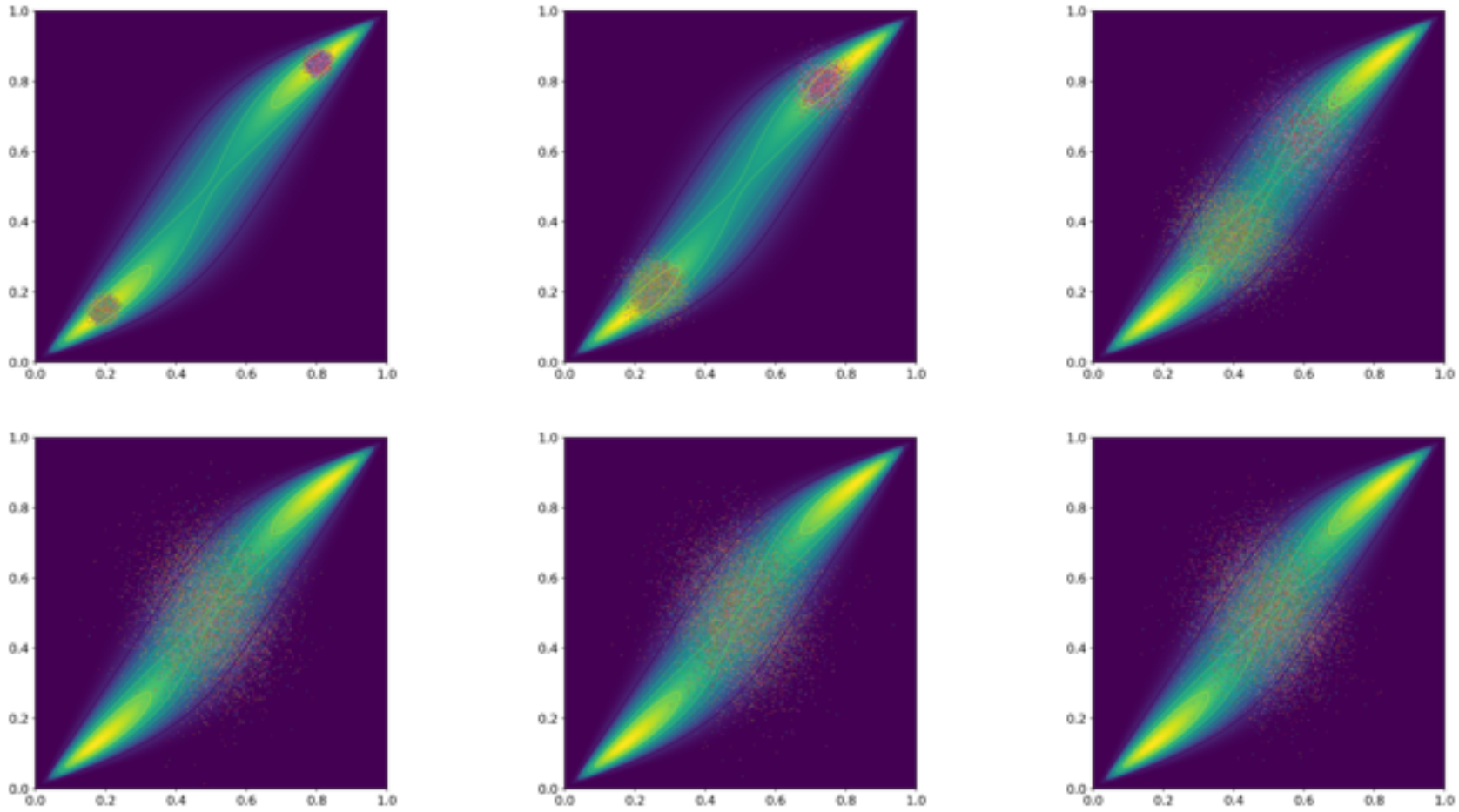

**Figure 4.2.3:** The evolution of the $A = 10$ single-component approximations after $1$k, $2$k, $3$k, $4$k, $5$k and $10$k training iterations, respectively. The component mean parameters are randomly initialized around $(0.5, 0.5)$ and individually maximize the ELBO objective to match the CoLN target distribution (its unnormalized density is shown via the contour plot in all figures; see Section 3.2 for details regarding CoLN). The colored scatter points are samples from the different components with separate colors for each component.

Moreover, even if initialization is not an issue, the global optima of the single-component objective, i.e. $\mathcal{L}_{\text{ELBO}}$, is likely different from maximizing $\mathcal{L}_{\text{MIS}}$ directly, and may even correspond to a single optimum although the target distribution multi-modal. Indeed, there appears to exist a single optimum—or, less stringently put, one very strong attractor—in the experiment setup corresponding to Figure 4.2.3.

Instead, as we showed in [2] (Paper B) it turns out that employing $\mathcal{L}_{\text{MIS}}$ in Equation (4.2.7) as an objective function for learning the variational parameters directly promotes diversity among the components [2]. In accordance with Definition 1, we call a variational distribution on the form of Equation (4.2.6) that maximizes the MISELBO objective a *variational mixture*.

### 4.3.1 The diversification mechanism in the MISELBO objective

There are at least two perspectives that explain the diversifying effect of the MISELBO objective. First, MISELBO can be decomposed into a cross-entropy and an

entropy term

$$\mathcal{L}_{\text{MIS}} = \frac{1}{A} \sum_{a=1}^{A} \mathbb{E}_{q_{\phi_a}(z|x)} \left[\log p_\theta(x, z_a)\right] + \mathbb{H}[q_{\phi_{1:A}}(z|x)], \tag{4.3.23}$$

and so maximization of the objective implies maximization of the entropy of the variational mixture. From information theory, we know that a larger entropy score typically corresponds to a flatter probability distribution, but when the distribution is a mixture, the entropy term also rewards separation of the mixture component densities. To see that mixture entropy maximization is a trade-off between diversity, i.e. component separation, and individual component flatness, we can study the following upper bound

$$\mathbb{H}[q_{\phi_{1:A}}(z|x)] = -\frac{1}{A} \sum_{a=1}^{A} \mathbb{E}_{q_{\phi_a}(z|x)} \left[\log \frac{1}{A} \sum_{a'=1}^{A} q_{\phi_{a'}}(z_a|x)\right] \tag{4.3.24}$$

$$\leq \frac{1}{A} \sum_{a=1}^{A} \mathbb{H}[q_{\phi_a}(z|x)] + \log(A). \tag{4.3.25}$$

The mixture entropy clearly promotes repulsion of the components, since the upper bound is obtained if $q_{\phi_{a'}}(z|x) = 0$ for all $a' \neq a$ when $z \sim q_{\phi_a}(z_a|x)$. However, after perfect separation (diversity) of the components, maximization instead implies maximization of the upper bound, and so the flatness of the individual components. Given fixed mixture weights (motivated below) and if the supports of the components are not disjoint by construction, then separation and flatness cannot be jointly maximized, which results in a trade-off.

In the adaptive importance sampling literature [55], it has been shown that the importance weights resulting from multiple importance samplers, i.e. mixtures of proposals, promote diversification. Specifically, set

$$w(z) = \frac{p_\theta(x, z)}{\frac{1}{A} \sum_{a=1}^{A} q_{\phi_a}(z|x)}, \tag{4.3.26}$$

then, assuming a fixed numerator, samples $z \sim q_{\phi_a}(z|x)$ that yield smaller $\sum_{a' \neq a} q_{\phi'_a}(z|x)$ will result in larger importance weights, $w(z)$. This guides the adaptation of the proposal parameters to produce diversely parameterized mixture components [55]. Identifying $w(z)$ as the ratio within the log-transform in Equation (4.2.7) makes the connection to MISELBO maximization clear.

Simultaneously, the cross-entropy term in Equation (4.3.23) attract the mixture components to high-probability regions. And so, collectively, the MISELBO objective results in variational mixtures that cooperate in order to explore the target posterior distribution.

**Motivating uniform mixture weights** In our works on variational mixtures we have been compelled to constrain ourselves to uniform mixture weights. This is mainly motivated by observations from two studies. In the adaptive importance sampling literature with multiple importance samplers it has been shown that uniform weights [49] promote exploration [55]. Meanwhile, in the BBVI literature [56], variational mixture inference with learnable mixture weights resulted in systematic mode collapse, which prevents the desirable cooperative behavior described above. In contrast to the findings in the latter study, we have found, in multiple works, that uniform weighting indeed produced diverse variational mixture components [2, 3, 57].

### 4.3.2 Related work

The first instance of variational mixture research seems to have been recorded 1997 in Jaakkola and Jordan [20],[2] where mixtures were motivated in order to improve upon single component mean-field approximations when the posterior is multi-modal. Later the same year, mixtures were applied to inference in sigmoid belief networks [19]. As closed-form optimization solutions for the ELBO are not available when the variational distribution is a mixture, the authors instead maximized a lower bound on the ELBO. Furthermore, a statement regarding potential performance gains made in this work was discussed above in Section 4.2.1.

**Variational mixtures in BBVI** As mentioned in the previous paragraph, variational mixtures in BBVI had indeed been considered prior to our work, namely in Morningstar et al. [56]. Our narratives bifurcate from the point that Morningstar et al. [56] conclude that the MISELBO objective (termed the stratified ELBO in their work) results in mode collapse when the mixture weights are learnable, which leads them to pursuit importance weighted ELBO variants (e.g. Equation (4.2.8)) as objective functions instead. The mode collapse behavior is expected when the weights are learnable in the sense that when one of the components end up in a high-probability region, an effective way to maximize the MISELBO objective is to simply assign this component a majority of the weight. As we showed in Kviman et al. [2], on the other hand, fixed uniform weighting of the mixture in the learning objective did not result in such a degeneration.

**Variational mixtures in CAVI** Slightly less related but of clear interest are the variational mixture works studied in the CAVI methodology. In Bishop and Nasrabadi [22, Ch. 10], and later Gershman, Hoffman, and Blei [53] and Blei, Kucukelbir, and McAuliffe [23], the CAVI update equations are made obtainable by lower-bounding the intractable mixture entropy, and so the cooperative behavior obtained from the MISELBO objective via the entropy term, as detailed in Section 4.3.1, is not enforced.

### 4.3.3 Experiments

We return to the density estimation task in Section 4.2.3, i.e. the task of minimizing the KL divergence to the two CoLN distributions in Figure 3.2.1. The experimental details are the same as in Section 4.2.3, except that we here use a variational mixture to approximate the target density.

[2] A preprint of Jaakkola and Jordan [20], with the title "*Approximating posteriors via mixture models*", was referenced with a to-appear citation in Bishop et al. [19].

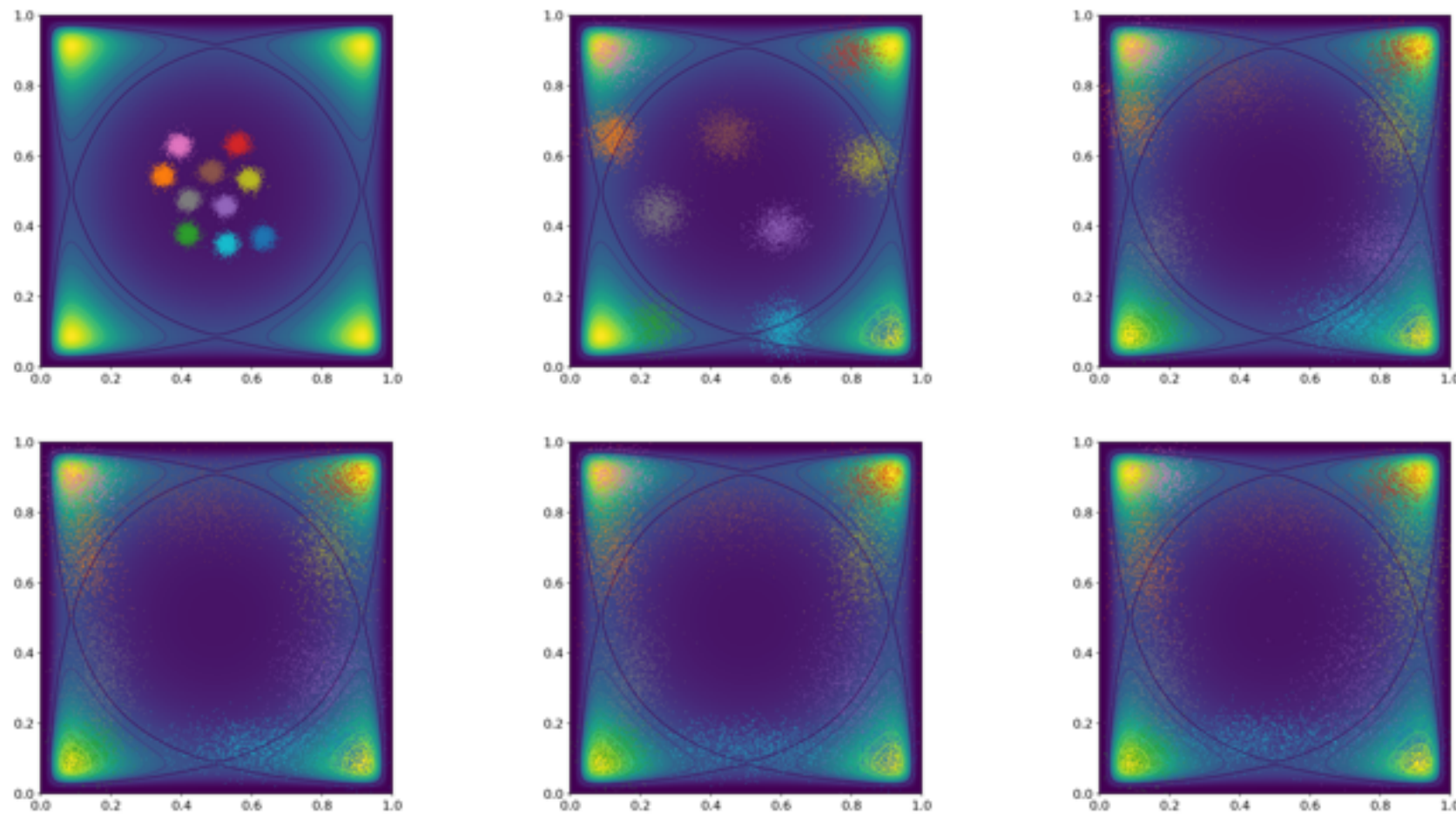

**Figure 4.3.4:** The evolution of the $A = 10$ variational mixture after $1$k, $2$k, $3$k, $4$k, $5$k and $10$k training iterations, respectively. The component mean parameters are randomly initialized around $(0.5, 0.5)$. The colored scatter points are samples from the different components with separate colors for each component.

In contrast to the single components in the ensemble approximations in Figure 4.2.1 and Figure 4.2.3, the variational mixture requires more steps before reaching the modes in quad-modal distribution, as shown in Figure 4.3.4. This is reasonably due to the cooperative nature of the variational mixture (Section 4.3.1), as the diversification mechanism constrains the components to cover the same mode and instead spread out.

The positive impact of the diversification mechanism on the approximation quality is clear from the curves in Figure 4.3.5, especially in the bi-modal setup: the gap between the two curves ($\Delta$ in Section 4.2.1) is consistently non-zero, meaning that the mixture has not degenerate as was the case for the ensemble approximation in Section 4.2.3. For both target distributions, the variational mixture MISELBO scores are better than those produced by the ensemble approximation in Figure 4.2.2—for reference, the ensemble approximation MISELBO scores were around $-12$ and $-2$ for the quad-modal and bi-modal targets, respectively.

Nonetheless, the MISELBO curves in Figure 4.3.5 show that the mixtures converge fast, while achieving higher (better) MISELBO scores than the ensemble approximation in Figure 4.2.2.

The superior performance of the variational mixture can also be verified through visual inspection of Figure 4.3.4 and Figure 4.3.6. It is especially pleasant to see how the components spread out within the contour plot of the bi-modal distribution (Figure 4.3.6) while maintaining a small mutual intersection in terms of support, especially when the component variances are small.

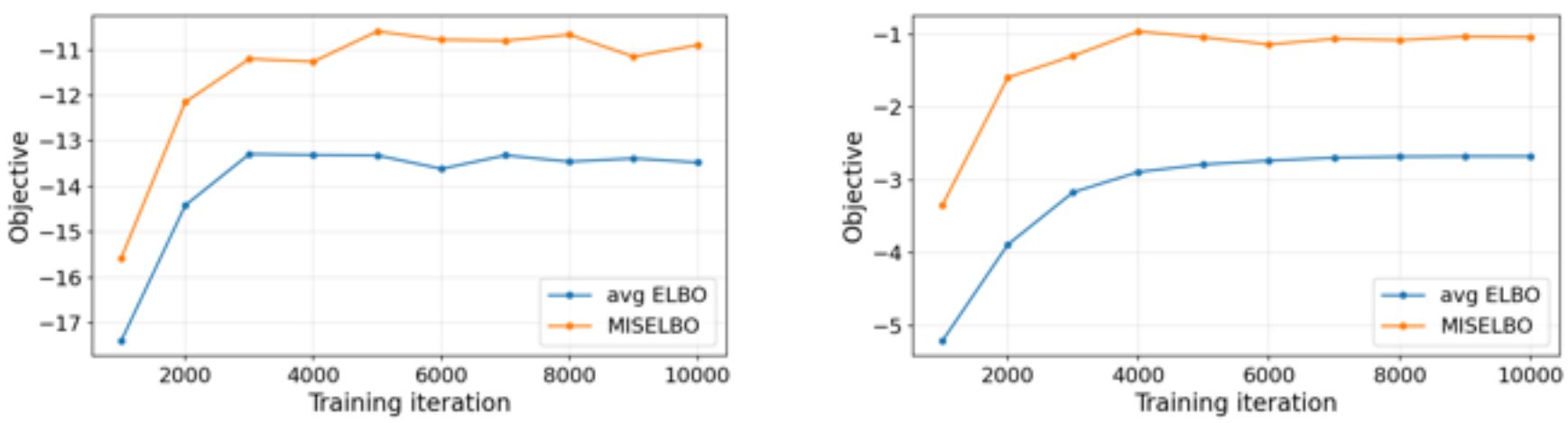

**Figure 4.3.5:** MISELBO and average (avg) ELBOs curves for the variational mixture on the quad-modal (left) and bi-modal (right) target densities.

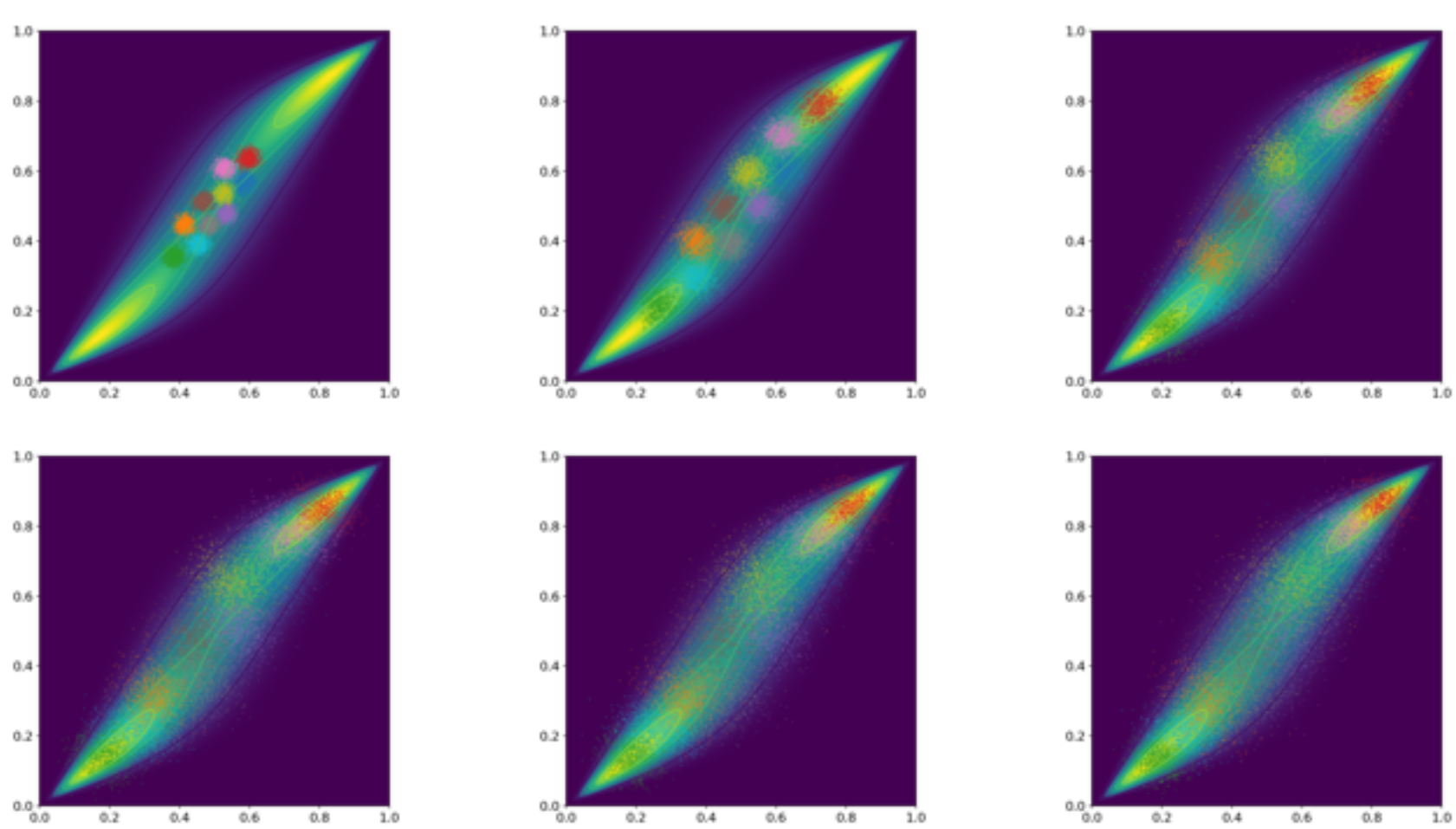

**Figure 4.3.6:** The evolution of the $A = 10$ variational mixture after $1$k, $2$k, $3$k, $4$k, $5$k and $10$k training iterations, respectively. The component mean parameters are randomly initialized around $(0.5, 0.5)$. The colored scatter points are samples from the different components with separate colors for each component.

## 4.4 Efficient variational mixture learning

As described in the related work section above, there have been various works on variational mixtures produced over a long time-span, but their potential to dominate the field of amortized approximate inference, demonstrated in Kviman et al. [2], has not yet been realized. One reason for this could be the early ideas about lower-bounding the mixture entropy in order to get closed-form inference equations [20, 22, 53, 23], which results in mixture learning schemes with reduced incentives to spread out the mixture components.

Another explanation of variational mixtures have not yet been widely adopted might be due to worries of inference and memory complexities growing with the number of mixture components. Let us revisit two statements regarding computational inefficiencies of variational mixtures found in the VI literature, starting with the one found in the seminal normalizing flows paper by Rezende and Mohamed [45]:

> "*[T]he mixture approach limits the potential scalability of variational inference since it requires evaluation of the log-likelihood and its gradients for each mixture component per parameter update, which is typically computationally expensive.*"

In adjacency to the above quote follows a reference to the work of Gershman et al. [53], where they introduce their non-parametric VI (NPV) algorithm and remark the following:

> "*Note that the number of parameters that need to be fit with NPV increases linearly with N (the number of components in the mixture). This may pose challenges for models with a large number of hidden variables.*"

These quotes both point to valid bottlenecks that limit the scaling of the number of variational mixture components, and in Kviman et al. [2] the best-performing algorithms do indeed make use of $A$ amortization networks—making models prohibitively large for high-dimensional datasets.

Furthermore, in BBVI algorithms for phylogenetic inference [44], the likelihood function does not naturally allow for automatic differentiation [58], which ultimately slows down the likelihood evaluation. This issue relates to the quote from [45]; inference of variational mixtures does not scale nicely with $A$ when the likelihood function is complicated to evaluate—although attempts have been made [57]. Beyond the likelihood evaluation, the mixture entropy term additionally requires an $\mathcal{O}(A \times A)$ computational complexity.

In our ICML 2024 paper [3] (Paper C), we tackled these two computational complexity issues. Concretely, the computational complexity of the MISELBO objective is reduced via the introduction of two new estimators in Section 4.4.1. In the subsequent Section 4.4.2 we show how to amortize the variational mixture inference in order to prevent a prohibitively rapid increase in the number of model parameters w.r.t. the number of mixture components.

### 4.4.1 Two new MISELBO estimators

To reduce the computational complexity inherent from the MISELBO objective function in Kviman et al. [2]—essentially the likelihood and the mixture entropy evaluations—we again found inspiration from the multiple importance sampling literature [59, 49]. Specifically, we derive two new efficient estimators of the MISELBO objective, some-to-all (S2A) and some-to-some (S2A).

Before we provide the exact formulations of the two new estimators, we categorize the standard estimator of the MISELBO objective as the all-to-all (A2A) estimator

$$\widetilde{\mathcal{L}}_{\text{A2A}} = \frac{1}{A}\sum_{a=1}^{A}\log\frac{p_\theta(x, z_a)}{\frac{1}{A}\sum_{a'=1} q_{\phi_{a'}}(z_a|x)}, \quad z_a \sim q_{\phi_a}(z|x) \tag{4.4.27}$$

That is, samples are drawn from all components, and then the denominator is computed for each of the samples, respectively. The estimating MISELBO costs $A$ likelihood computations and $A \times A$ density evaluations to compute the denominator.

If the likelihood calculation is expensive (e.g. in the phylogenetics example above, or if the likelihood parameters are given by a large neural network with expensive forward passes), then we can subsample $S$ component indices without replacement, and then generate latents from the $S$ components but keep $A$ components in the denominator, i.e. some-to-all

$$\widetilde{\mathcal{L}}_{\text{S2A}} = \frac{1}{S}\sum_{s=1}^{S}\log\frac{p_\theta(x, z_s)}{\frac{1}{A}\sum_{a=1} q_{\phi_a}(z_s|x)}, \quad z_s \sim q_{\phi_s}(z|x). \tag{4.4.28}$$

This estimator is an unbiased estimator of MISELBO [3], but at the cost of keeping the $\mathcal{O}(A \times A)$ complexity of estimating the mixture entropy. On the other hand, we have now reduced the number of likelihood computations to $S < A$.

The second, new estimator sacrifices unbiasedness in order to avoid the $\mathcal{O}(A \times A)$ mixture entropy cost, namely by keeping only the subsampled set of $S$ components in denominator,

$$\widetilde{\mathcal{L}}_{\text{S2S}} = \frac{1}{S}\sum_{s=1}^{S}\log\frac{p_\theta(x, z_s)}{\frac{1}{S}\sum_{s'=1} q_{\phi_{s'}}(z_s|x)}, \quad z_s \sim q_{\phi_s}(z|x). \tag{4.4.29}$$

As said, the S2S estimator is a biased estimator of MISELBO, and the number of likelihood evaluations is still $S$, but the estimator enjoys a substantial reduction in mixture entropy estimation complexity, now $\mathcal{O}(S \times S)$. This allows us to pick a very large $A$, while controlling the computational cost through $S$. Notably, if we choose $S = 1$ and $A > 1$, then maximizing the MISELBO objective with the S2S estimator is equivalent to inferring single-component approximations in an ensemble with $A$ components.

### 4.4.2 Amortized variational mixture inference

The number of amortization networks (or encoders in the VAE terminology) in Kviman et al. [2] increased linearly with $A$. A linear increase in network (or model) parameters is indeed difficult to avoid, however we can be clever about the construction of the amortization network such that the gradient of the parameter increase has smaller magnitude.

In Hotti et al. [3], we introduce a new amortization scheme specifically designed to allow for more efficient scaling of the network parameters in mixture learning. Let $O_A(s) \in \{0,1\}^A$ denote an $A$ long one-hot encoding with $s$ as the non-zero entry. Then let $H : \mathcal{X} \mapsto \mathbb{R}^d$ be a neural network and denote $d$ as the number of dimensions in some auxiliary space. The network $H$ is shared across all mixture components.

Given a data point $x$, mapped to the auxiliary space via $H$, concatenate $H(x)$ with $O_A(s)$ to get the $d+A$ long $h$ vector. Finally, let $F : \mathbb{R}^{d+A} \mapsto \Phi$ denote a second neural network, where $\Phi$ denotes the space of variational parameters. Fixing $x$ and iterating over $s$ entries in $O_A(s)$ will hence produce the parameters of the variational mixture, i.e.,

$$\phi_s = F\left([H(x), O_A(s)]^\top\right), \quad \forall s \in [1, A]. \tag{4.4.30}$$

The linear growth in parameters with $A$ is now constrained to the linear growth of the input dimensions of $H$. The network $H$ can be chosen to be small, which makes amortized variational mixture inference a neat, practical way of dramatically decreasing in the number of network parameters [3].

### 4.4.3 Experiments

There is no need for amortization in the recurring two-dimensional CoLN density estimation task—one could introduce a more advanced parameterization of the Gaussian components using a neural network, but this is an excessive alternative in this case. Instead, see [3] for a demonstration of the power of amortizing the variational mixture inference.

Furthermore, the S2A estimator in Equation (4.4.28) is a highly recommended alternative to the costly default A2A estimator in Equation (4.4.27) when the target distribution is expensive to compute. The CoLN distribution is not expensive, however, especially in 2D, and so it is more interesting to focus on the properties of the S2S estimator here.

To utilize the efficiency of the S2S estimator in an interesting way, we double the number of components in the mixture compared to before (i.e., $A = 20$ now) and set $S = 10$. Thus we ensure that the S2S estimator has the same time complexity as a $10$ component variational mixture with the A2A estimator (as in Section 4.3.3).

Similarly as for a variational mixture trained with the A2A estimator, the mixture components in Figure 4.4.7 avoid support overlap when exploring low-probability regions (the darker regions are assigned less density by the CoLN distribution). The S2S-based approximation requires more training iterations to convergence than the A2A estimator (left plot in Figure 4.4.8) but allows for scaling up the number of mixture components without increased cost in computational complexity. The same analysis holds for the bi-modal setting in Figure 4.4.8 and Figure 4.4.9.

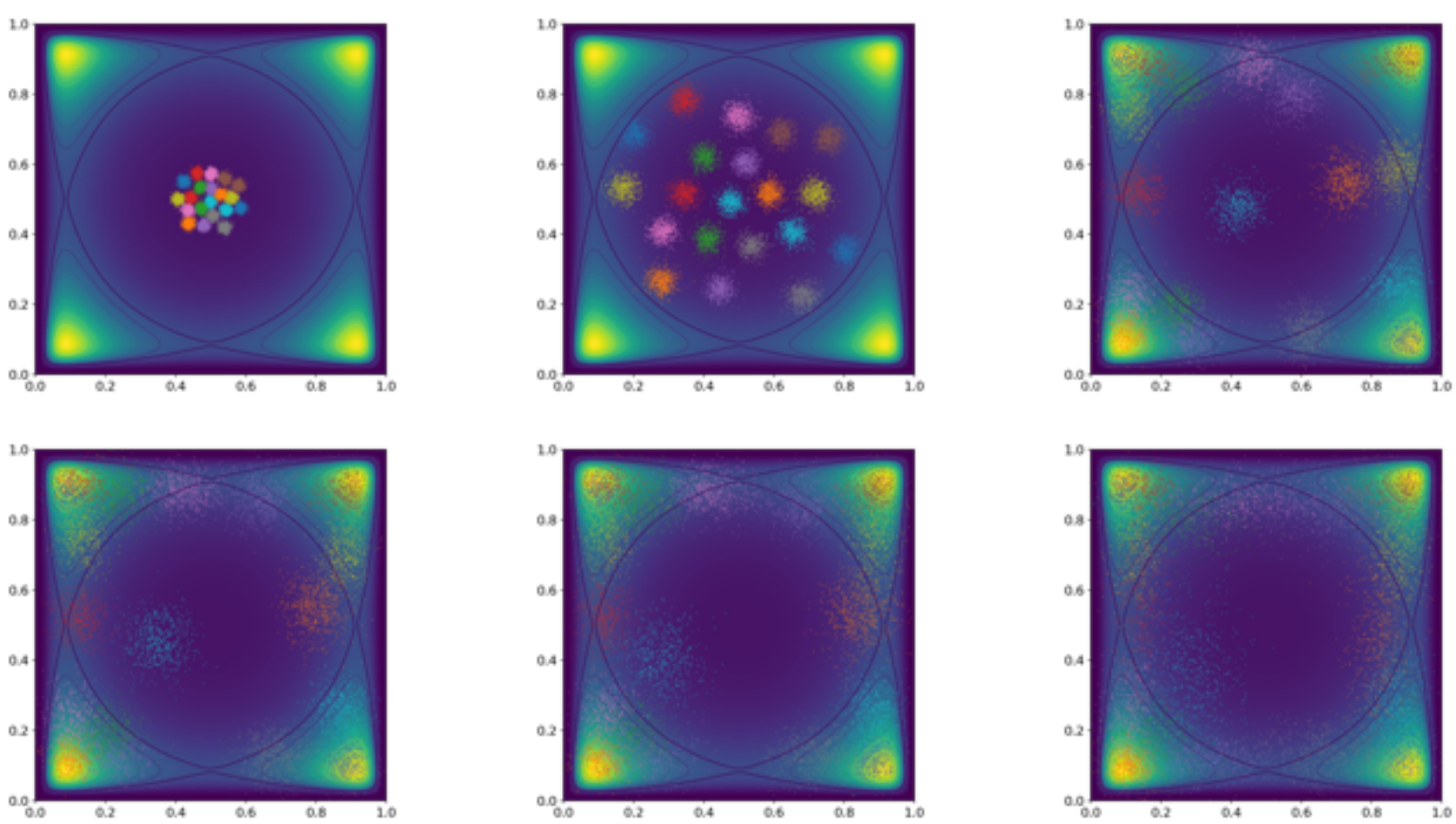

**Figure 4.4.7:** The evolution of an $A = 20$ variational mixture trained with the S2S estimator ($S = 10$) after $1$k, $5$k, $8$k, $9$k, $10$k and $20$k training iterations, respectively. The colored scatter points are samples from the different components with separate colors for each component.

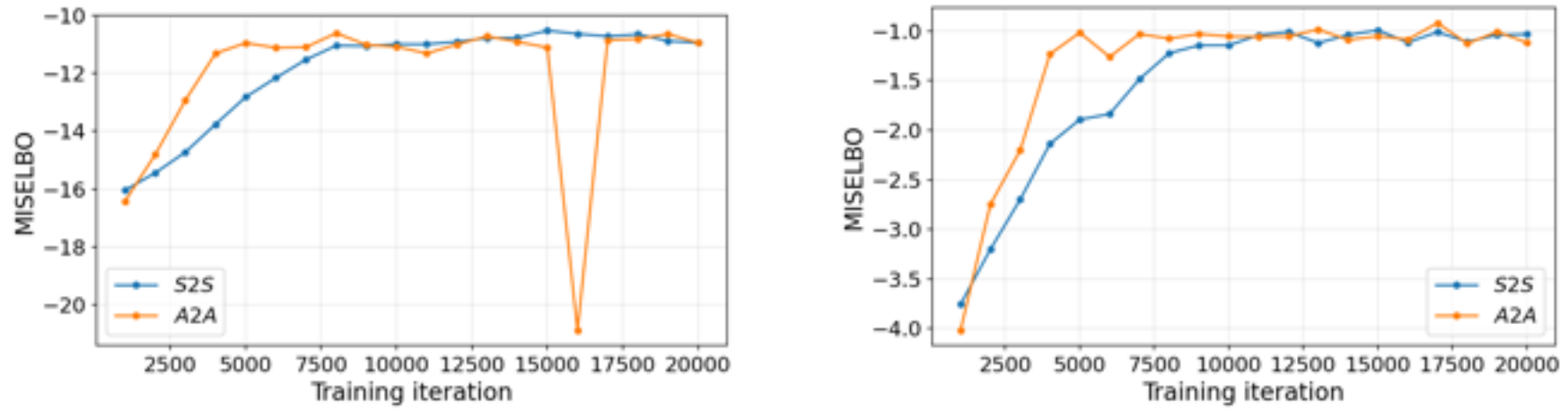

**Figure 4.4.8:** MISELBO curves for a $20$ component variational mixture learned with the S2S estimator vs. a $10$ component variational mixture with the A2A estimator on the quad-modal (left) and bi-modal (right) target densities. The time complexities of the two different inference algorithms are equal.

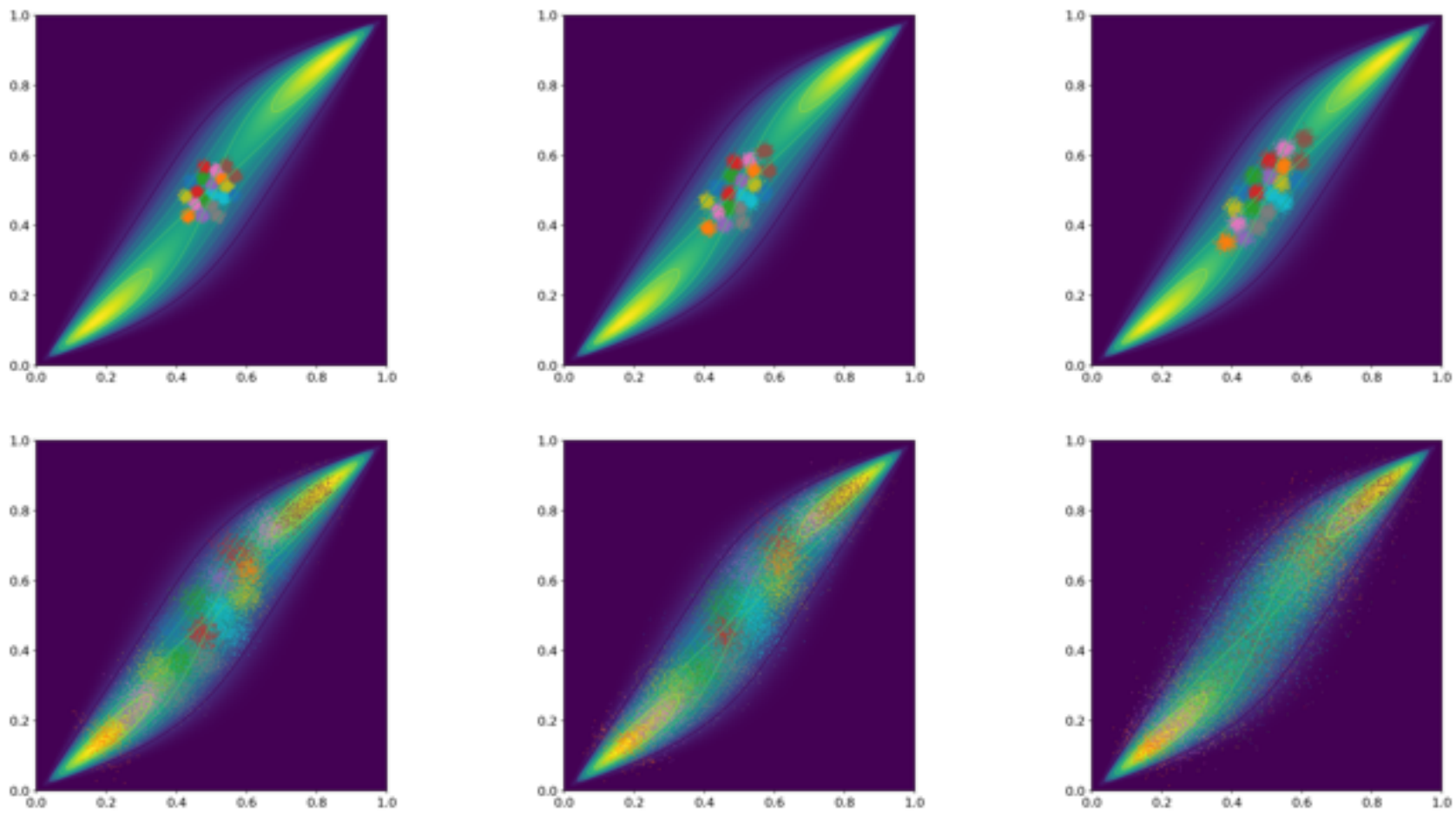

**Figure 4.4.9:** The evolution of an $A = 20$ variational mixture trained with the S2S estimator ($S = 10$) after $1$k, $2$k, $3$k, $8$k, $10$k and $20$k training iterations, respectively. The component mean parameters are randomly initialized around $(0.5, 0.5)$. The colored scatter points are samples from the different components with separate colors for each component.

In this experimental setup, doubling the number of mixture components seems to have only marginal benefits in terms of approximation accuracy (Figure 4.4.8), but one can of course easily construct scenarios where increasing the number of mixture components has large impact on the density estimation accuracy. For example, in Figure 4.4.10 we compare two-component variational mixture inferred with the A2A estimator (top left), and a variational mixture inferred via the S2S estimator with $S = 2$ but which disposes of four components $A = 4$ (top right). This design ensures that the S2S-based mixture is inferred at the same computational cost as the A2A alternative, but with clear approximate improvements, both visually and numerically (bottom plot in Figure 4.4.10). We previously demonstrated the benefit of the S2S estimator in Hotti et al. [3, Figure 5], scaling up to $A = 50$ on the MNIST dataset with superior marginal log-likelihood scores *and* lower computational cost than an A2A-based baseline.

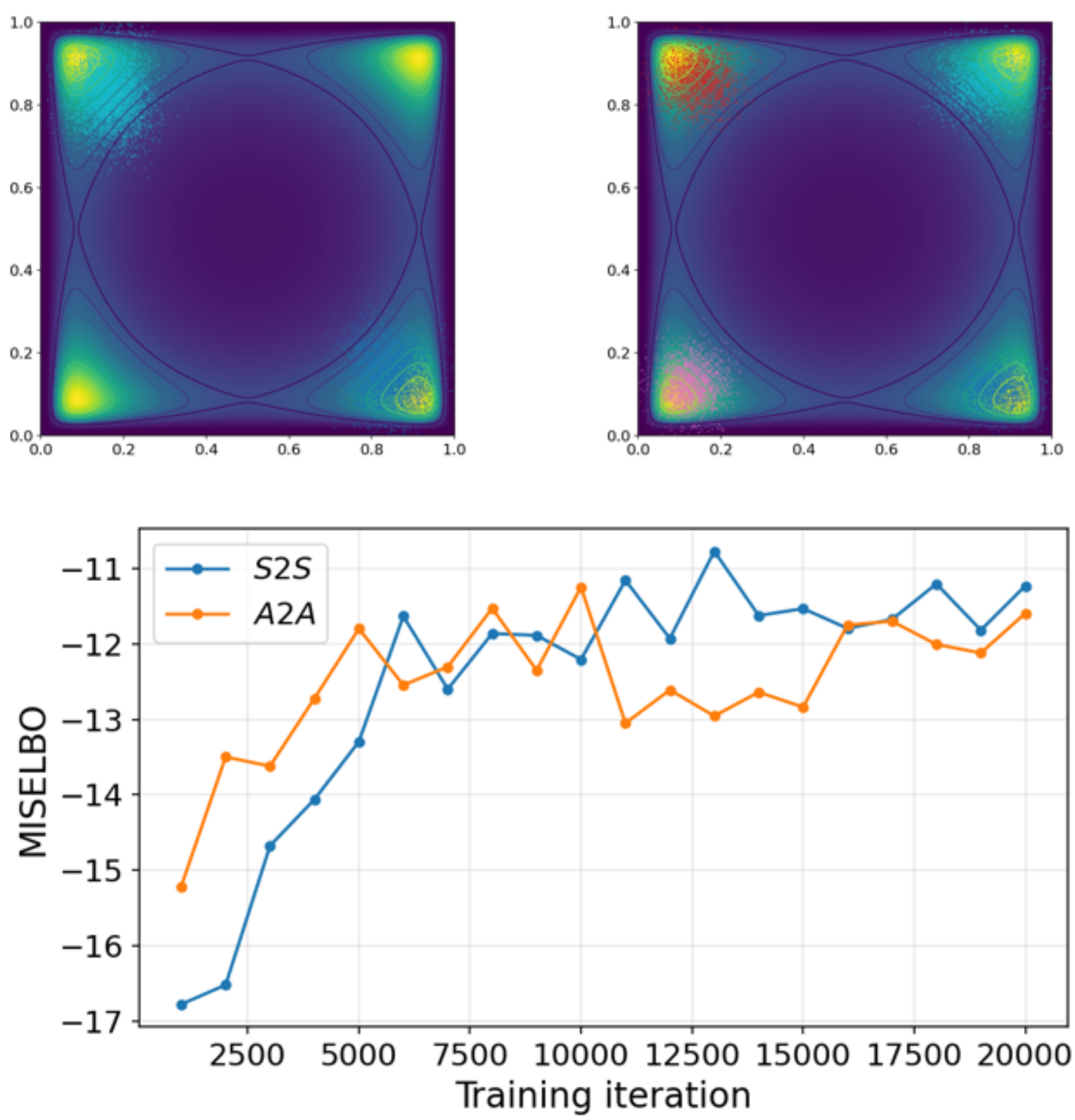

**Figure 4.4.10:** The variational approximations of the quad-modal CoLN distribution using a two-component mixture inferred with the A2A estimator (top left) vs. one with four components inferred via the S2S estimator (top right; $S = 2$, $A = 4$). There is no increase in computational complexity between the two alternatives, but there is a clear visual and numerical (bottom) justification for using the S2S estimator. The colored scatter points are samples from the different components with separate colors for each component, and the shown approximations are the ones obtained after $20$k training iterations.

# 5 Flow Matching

This chapter continues the methodological trajectory of the thesis by moving from variational inference to flow matching [46], a recently developed framework for learning deterministic vector fields whose induced dynamics transform one probability distribution into another.

After reviewing the flow matching formulation in the standard bi-marginal setup, we describe the necessity of specialized methods when moving to multi-marginal problems, and we explain how Paper D in this thesis [4] is purposed to solve these challenges.

Finally, by combining the methods and insights from all contributed papers, Paper A-D [1, 2, 3, 4], we derive a new multi-marginal flow matching method. This new generative model serves as a unification of the methodological themes developed throughout the thesis and constitutes an exciting direction for future advancement of the field.

## 5.1 A bridge between variational inference and flow matching

In Section 4.1, we introduced NFs and categorized them as a BBVI tool for constructing more sophisticated posterior approximations. While this is not incorrect—after all, the title of the seminal NF article [45] is "*Variational inference with normalizing flows*"—the methodology eventually started to cultivate outside the approximate Bayesian inference field [60], and in Chen et al. [61], a useful reformulation of NFs was provided:

NFs are composed of a series of transformations, iteratively pushing a simple distribution, like a Gaussian, to a complex distribution that, hopefully, approximates a target distribution more accurately than the simple distribution. The key insight from [61] to re-interpret NFs, is that these sequences can be seen as discretizations of a continuous-time function.

Specifically, if we regard the discretization as an operation along a "*time*" axis, then, in the continuous-time limit (the number of discretization steps go to infinity), the discretized function is a vector field in an ordinary differential equation (ODE) that describes the time-dependent dynamics of transporting one distribution to another along the time axis $t \in [0, 1]$ [61]. Continuous NFs (CNFs) are thus models where ODEs are simulated, and where the vector field (a neural network) is trained via likelihood-based maximization. Unfortunately, ODE simulations during training along with an expensive objective made CNFs slow to train.

## 5.2 Bi-marginal flow matching

Flow matching (FM; [46]) was proposed as a simulation-free alternative for learning CNFs, resulting in substantial reductions in training time. In FM, the CNF learning task is cast as the following inference task: given $n$-dimensional samples from two marginal distributions at times $t = 0$ and $t = 1$, i.e. $x_0 \sim \nu_0$ and $x_1 \sim \nu_1$, learn a vector field, $u_t^\theta : \mathcal{X} \times [0, 1] \mapsto \mathbb{R}^n$ (the CNF) that approximates a target vector field $v_t$ which generates the dynamics of transporting samples from $\nu_0$ to $\nu_1$, described by an ODE with a stochastic initial condition

$$\frac{d\psi_t(x_0)}{dt} = v_t(\psi_t(x_0)) \tag{5.2.1}$$

$$\psi_0(x_0) = x_0, \quad x_0 \sim p_0, \tag{5.2.2}$$

where $\psi_t : \mathcal{X} \times [0, 1] \mapsto \mathcal{X}$ is the flow, i.e. a mapping from $x_0$ to $x_t$, $x_t = \psi_t(x_0)$. For a stochastic initial condition $x_0 \sim p_0$, the flow induces a probability path $p_t = (\phi_t(\cdot))_\# p_0$.

If we set $\nu_0 = p_0$ and $\nu_1 = p_1$, and if $v_t$ is accurately approximated by $u_t^\theta$, then simulating an ODE parameterized by $u_t^\theta$ will generate samples distributed as $(\psi_t(\cdot))_\# \nu_0$, with

$$\nu_0 = (\psi_0(\cdot))_\# \nu_0 \tag{5.2.3}$$

$$\nu_1 = (\psi_1(\cdot))_\# \nu_0. \tag{5.2.4}$$

This is a powerful framework for learning CNFs in a simulation-free manner as learning of the vector field is achievable via a simple regression loss, the FM objective:

$$\mathcal{L}_{\text{FM}} = \mathbb{E}_{t,x_t} \left[ \left\| u_t^\theta(x_t) - \frac{dx_t}{dt} \right\|^2 \right], \quad t \sim U[0, 1], x_t = \psi_t(x_0), x_0 \sim \nu_0, \tag{5.2.5}$$

where $dx_t/dt = d\psi_t(x_0)/dt = v_t(\psi_t(x_0))$ (Equation (5.2.1)). Upon successful minimization of the FM loss above, $u_t^\theta$ generates the probability path induced by the target vector field, $p_t$. The possibility to generate $p_t$ is essential in order to solve the $u_t^\theta$-ODE for arbitrary initial conditions.

Due to intractability, we cannot compute the flow or the vector field, and thus we cannot sample from $p_t$. To understand where the intractability arises, write the exact expression of the time-dependent marginal distribution as

$$p_t(x) = \int_{(x_0,x_1)} p_t(x|x_0,x_1)p_0(x_0)p_1(x_1)dx_0x_1 = \int_{(x_0,x_1)} p_t(x|x_0,x_1)d\nu_0 d\nu_1, \quad (5.2.6)$$

and note that integration over the end marginals is intractable as the integral is over a high-dimensional integrand, and since $p_t(x)$ becomes an infinite mixture with no known closed-form for general $\nu_0$ and $\nu_1$. The intractability can also be grasped from the following rhetorical question: how should we design a flow $\psi_t$ that maps a point $x_0$ to the support of $\nu_1$ in order to satisfy Equation (5.2.4)?

Interestingly, a new object pops up in Equation (5.2.6), namely the conditional probability path, $p_t(x|x_0,x_1)$. Fortunately, it turns out that learning to generate this conditional probability path is both tractable and equivalent to learning to generate the marginal probability path [46].

**Conditional flow matching** Let $\pi : \mathcal{X} \times \mathcal{X} \mapsto \mathcal{X}$ denote a joint distribution over bi-marginal sample pairs $(x_0, x_1)$. This probability distribution is typically called a coupling, and the default coupling choice assumes an independent sampling procedure, referred to as the independent coupling $\pi(x_0,x_1) = \nu_0\nu_1$ [62].

The coupling is a probabilistic approach to pairing $(x_0,x_1)$ pairs, which makes the inference task tractable by allowing a conditional flow $\psi_t(x_0,x_1) : \mathcal{X} \times \mathcal{X} \times [0,1] \mapsto \mathcal{X}$ which induces a conditional vector field

$$\frac{dx_t}{dt} = v_t(x_t|x_0,x_1), \quad x_t = \psi_t(x_0,x_1) \quad (5.2.7)$$

that generates the conditional probability path, $p_t(x|x_0,x_1) = (\psi_t(\cdot,\cdot))_{\#}\pi$.

We can plug in the conditional versions of the relevant objects in Equation (5.2.5) and marginalize over $(x_0,x_1)$ pairs via the coupling to get the conditional flow matching (CFM),

$$\mathcal{L}_{\text{CFM}} = \mathbb{E}_{t,x_t}\left[\left\|u_t^\theta(x_t) - \frac{dx_t}{dt}\right\|^2\right], \quad t \sim U[0,1], x_t = \psi_t(x_0,x_1), (x_0,x_1) \sim \pi. \quad (5.2.8)$$

State-of-the-art examples of conditional flows, a.k.a. interpolants, includes the linear interpolant [46]

$$\psi_t(x_0,x_1) = x_t = tx_1 + (1-t)x_0 \quad (5.2.9)$$

which leads to the following conditional vector field

$$\frac{dx_t}{dt} = x_1 - x_0, \quad (5.2.10)$$

as well as conditional flows that avoid Euclidean geometry assumptions via Riemannian metric-based geodesics [63].

One can also flip the modeling order and directly choose a conditional probability path that, inversely, gives a conditional flow. For instance, one can consider a Gaussian probability path [46]

$$p_t(x|x_0, x_1) = \mathcal{N}\left((\mu_t(x_0, x_1), \sigma^2(x_0, x_1)\right), \tag{5.2.11}$$

which admits a conditional flow via the reparameterization trick (see Section 4.1 or [40]),

$$\psi_t(x_0, x_1) = \mu_t(x_0, x_1) + x\sigma^2(x_0, x_1), \quad x \sim \mathcal{N}(0, I_n), \tag{5.2.12}$$

and consequently a unique target vector field [46, Theorem 3]

$$v_t(x|x_0, x_t) = \frac{d}{dt}\psi_t(x_0, x_1) = \frac{x - \mu_t(x_0, x_1)}{\sigma_t(x_0, x_1)}\frac{d}{dt}\sigma_t(x_0, x_1) + \frac{d}{d_t}\mu_t(x_0, x_1). \tag{5.2.13}$$

## 5.3 Multi-marginal flow matching

In multi-marginal flow matching (MMFM) we consider a collection of $K > 2$ marginal distributions

$$x_{t_i} \sim \nu_{t_i}, \qquad 0 = t_1 < t_2 < \cdots < t_K = 1, \tag{5.3.14}$$

where samples are unpaired across time, and study how to do interpolation between the marginal samples. Two potential challenges are immediately apparent:

- Pointwise interpolations between samples from consecutive marginals may cause “kinks” in the conditional flow.
- The geometry of the data may change with time.

The first point is a concern as kinks will make the target vector field change sign rapidly w.r.t. $t$, regardless how we choose to construct the pointwise interpolation. This in turn will result in high variance in the CFM loss [4].

The second point is a concern if the interpolation is pointwise, i.e. for conditional flows that satisfy

$$\psi_t(x_{t_{i-1}}, x_{t_i}) = \begin{cases} x_{t_{i-1}}, & t = t_{i-1} \\ x_{t_i}, & t = t_i \end{cases} \tag{5.3.15}$$

for all $i \in [2, K]$ [62, 64, 65, 66], or if the conditional flow is a projection onto a time-independent data manifold [63].

In our work in Kviman et al. [4] (Paper D), we circumvent both of the challenges in the bullet list above, which we explain next.

## 5.4 Learning interpolants in MMFM with an adversarial objective

Let $\pi$ be a coupling between $\nu_0$ and $\nu_1$. For each pair $(x_0, x_1) \sim \pi$, we define a family of interpolants

$$G_\phi(x_0, x_1, t) = (1-t)x_0 + tx_1 + t(1-t)f_\phi(x_0, x_1, t), \tag{5.4.16}$$

where $f_\phi$ is a neural network parameterizing deviations from linear interpolation. This parametrization guarantees correct end-marginals at $t = 0$ and $t = 1$, and is inspired by the formulations in Kapusniak et al. [63] and Neklyudov et al. [64]. An important distinction from the previous works is that our interpolation is always parameterized by samples from the two end-marginals, $\nu_0$ and $\nu_1$, which is not the case for the related works where the linear interpolation in the parameterization is designed between samples from consecutive marginal distributions.

Crucially, we do not require interpolants to pass through individual observed samples, but only that their pushforward distributions match the target marginals. As such, we impose a *distributional* constraint: for each intermediate time $t_i$,

$$(G_\phi(\cdot, \cdot, t_i))_\# \pi = \nu_{t_i}. \tag{5.4.17}$$

We enforce this condition via an adversarial objective, interpreting the interpolant as a conditional generator whose "noise" is the pair $(x_0, x_1) \sim \pi$. A discriminator is trained to distinguish samples from $\nu_{t_i}$ and interpolated samples (i.e. Equation (5.4.16)) at $t_i$. Minimizing the resulting generative adversarial net (GAN [67]) objective corresponds to minimizing a JSD [67] between the two distributions $(G_\phi(\cdot, \cdot, t_i))_\# \pi$ and $\nu_{t_i}$. In other words, let the discriminator $D_\gamma : \mathcal{X} \times [0,1] \mapsto [0,1]$ be a neural network with parameters $\gamma$, then the adversarial objective is

$$\min_{G_\phi} \max_{D_\gamma} \mathbb{E}_{(x_0,x_1)\sim\pi} \left[\log\left(1 - D_\gamma(G_\phi(x_0, x_1, t), t)\right)\right] + \mathbb{E}_{x_t \sim \nu_t} \left[\log D_\gamma(x_t, t)\right], \tag{5.4.18}$$

which we marginalize over all $t \in \{t_i\}_{i=2}^{K-1}$. Note that this is amortized inference (see Section 4.1) w.r.t. $t$ which results in a time-dependent generator with pushforward according to Equation (5.4.17).

The deviation of our approach to the existing MMFM methods is fundamental, as they typically enforce pointwise constraints, for instance by spline interpolation [65, 66] through samples or by metric-based geodesics [63]. In contrast, our formulation matches intermediate marginals in distribution.

Distributional matching makes the interpolant free to rearrange mass within a marginal, provided that its pushforward aligns with the observed distribution. As such, there are infinitely many solutions to the inference problem. This is an undesirable feature as it does not guarantee that $(x_0, x_1)$ samples give interpolations $G_\phi(x_0, x_1, t)$ that are close to $(x_0, x_1)$. Informally, the correction network $f_\phi$ can, theoretically, go arbitrarily crazy in order to match the marginal distributions.

To constrain the flexibility of the generator, we incorporate regularization terms that penalize excessive curvature or deviations from suitable reference interpolants.

Under mild absolute continuity assumptions, these regularizers admit uniqueness guarantees at each time slice, providing theoretical grounding for the learned interpolants. For completeness, the regularizers with the linear and piecewise linear reference interpolants are described next.

Denote the linear reference interpolant as

$$\ell(x_0, x_1, t) = (1-t)x_0 + tx_1, \tag{5.4.19}$$

then the regularizing term, added as a constraint to Equation (5.4.18) with weight $\lambda$, is

$$L_{\text{reg}}(G_\phi, t_i) = \mathbb{E}_{(x_0,x_1)\sim\pi}\left[\|G_\phi(x_0, x_1, t_i) - \ell(x_0, x_1, t)\|_2^2\right], \tag{5.4.20}$$

and guarantees that the solution to the regularized version of the adversarial objective Equation (5.4.18) is unique [4]:

**Theorem 3.** *Fix $t \in (0,1)$, $\nu_t$ and a coupling $\pi$ such that $(\ell(\cdot,\cdot,t))_{\#}\pi$ i absolutely continuous w.r.t. the Lebesgue measure. Then the interpolant $G_\phi(\cdot,\cdot,t)$ that minimizes Equation* (5.4.20)*, subject to $(G_\phi(\cdot,\cdot,t))_{\#}\nu_0 = \nu_t$, exists and is unique.*

The linear reference may be overly restrictive if the dynamics are changing substantially with time, and so we also consider the slightly more data driven, piecewise linear interpolant

$$\ell(x_0, x_1, x_{t_i}, t) = \begin{cases} \frac{tx_{t_i} + (t_k - t)x_0}{t_i}, & t \leq t_i, \\ \frac{tx_1 + (1-t)x_{t_i}}{1-t_i}, & t > t_i, \end{cases} \tag{5.4.21}$$

and the corresponding regularizer that regresses $G_\phi$ to $\ell(x_0, x_1, x_{t_i}, t)$, marginalized over $t \in [0,1]$,

$$L_{\text{reg}}(\phi; t_i) = \mathbb{E}_{t\sim U[0,1]}\mathbb{E}_{(x_1,x_{t_i},x_0)\sim\pi_{t_i}}\left[\|G_\phi(x_0, x_1, t) - \ell(x_0, x_1, x_{t_i}, t)\|^2\right], \tag{5.4.22}$$

where we let $\pi_{t_i}$ be a Markov-chained OT coupling [62] $\pi_{t_i} = \pi(x_1|x_{t_i})\pi(x_{t_i}|x_0)\nu_0(x_0)$. This alternative regularization scheme also admit a uniqueness guarantee [4]:

**Theorem 4.** *Fix $t \in (0,1)$, $\nu_t$ and a Markov-chained coupling $\pi_{t_i} = \pi(x_1|x_{t_i})\pi(x_{t_i}|x_0)\nu_0(x_0)$, and let $\tilde{\pi}$ denote $\pi_{t_i}$ marginalized over all $(x_0, x_1)$. Set $\ell(x_0, x_1, x_{t_i}, t)$ as in Equation* (5.4.21)*, and assume that $(\ell(\cdot,\cdot,\cdot,t))_{\#}\pi_{t_i}$ is absolutely continuous w.r.t. the Lebesgue measure. Then the interpolant $G_\phi(\cdot,\cdot,t)$ that minimizes Equation* (5.4.22)*, subject to $(G_\phi(\cdot,\cdot,t))_{\#}\tilde{\pi} = \nu_t$, exists, is unique and is a minimizer for all $t \in [0,1]$.*

See Kviman et al. [4, Appendix A] for the proofs. We named the interpolants that are trained via this regularized adversarial loss *adversarially learned interpolants* (ALIs). Once trained, the ALIs are marginalized using the CFM loss Equation (5.2.8), yielding a globally defined vector field (ALI-CFM) whose induced marginals agree with all observed time stamps.

We showed via experimentation with synthetic and real data that our distributional fitting approach resulted in vector fields that more accurately modeled the underlying dynamics [4].

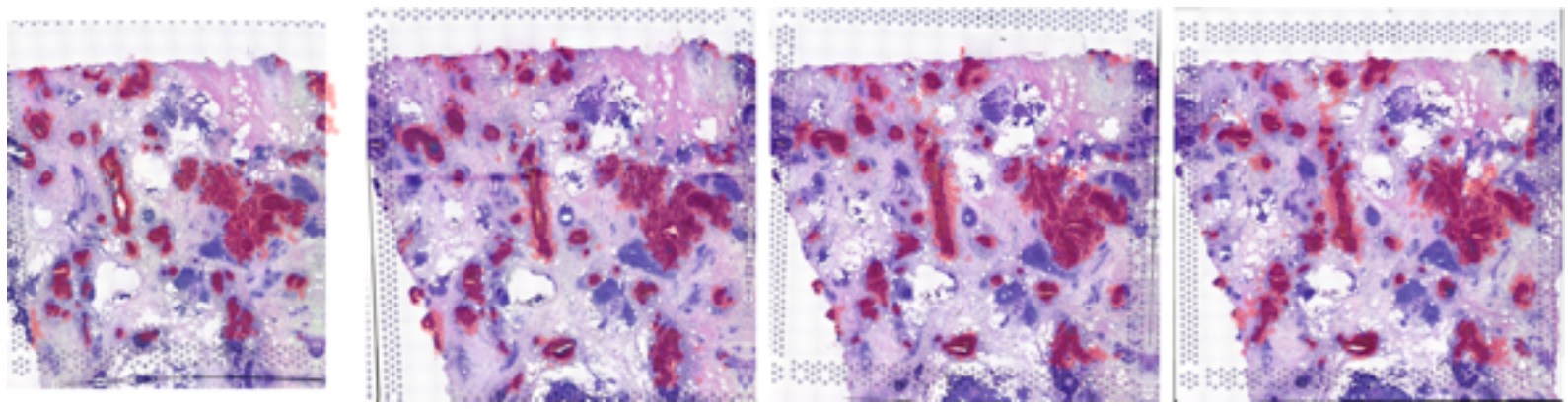

Figure 5.4.1: Preprocessed 3D-stacked ST breast cancer data. The H&E-stained images are overlaid with red scatter points that represent coordinates spots that have been assigned a tumor purity score greater than $0.8$ according to the raw data in Mo et al. [9].

### 5.4.1 3D spatial transcriptomics

There is one experiment in Kviman et al. [4] that deserves special attention—not only due to the superior performance of our ALI-CFM, but also due to the novelty of the experimental design—namely the reconstruction of a three-dimensional (3D) tumor geometry from two-dimensional spatial transcriptomics (ST) sections. See Section 2.2.2 for a background on ST.

**Problem formulation.** In our 3D ST problem setting, we have access to sequence of tissue slices, such that for each section $i = 1, \dots, K$, we observe $x_{t_i} \sim \nu_{t_i}$, where the sample $x_{t_i}$ is a tumor-annotated spatial spot coordinate (a red scatter point in Figure 5.4.1). Note that although the subscripts above have "time" subscripts, there is no actual temporal aspect to the data, but the sections are assumed to be stacked along a spatial $z$ axis. We stick to the $t$ notation for consistency. The slices are obtained through physical sectioning of a tumor volume, and therefore represent noisy, misaligned, and partially distorted observations of an underlying three-dimensional structure.

The inference task is to predict coordinates from a held-out marginal dataset using the observed marginal samples, thereby modeling how the spatial geometry evolves across slices. The geometry of the tumor volumes can exhibit branching or looping structures, as illustrated in Figure 5.4.2, resulting in inherently multi-modal intermediate distributions.

**Alignment preprocessing.** We worked with a multi-slide ST breast cancer dataset from [9]. To ensure that the spatial coordinates of each slice are expressed in a common reference frame, we aligned the data as part of a preprocessing step. As discussed in Kviman et al. [4, Appendix C], the raw H&E-stained images were first pairwise aligned using nonlinear thin-plate spline transformations. The learned pixel-wise warps were subsequently applied to the spot coordinates, yielding a sequence of aligned point clouds. The labeling of the coordinates as tumorous was made based on a tumor purity score available in the dataset [9]. The range of the purity score was $[0, 1]$, and we categorized all coordinates with scores $\geq 0.8$ as tumor coordinates.

This preprocessing step is crucial: without alignment, the marginals $\nu_{t_i}$ would differ not only due to biological variation, but also due to extrinsic rotations, scalings, and local deformations introduced during sectioning.

**What makes the 3D ST problem challenging?** An issue that has frequently been raised regarding GANs is that they are challenging to train when the target distribution is multi-modal. Even after alignment, the intermediate marginals remain highly multi-modal which indeed initially caused mode-collapse issues for ALI-CFM: using a bi-marginal OT-coupling $\pi(x_0, x_1)$, and uniform sampling of $x_{t_i} \sim \nu_{t_i}$ in Equation (5.4.18), the linear reference trajectory that connects $(x_0, x_1)$ does likely not pass through a point at $t = t_i$ close to $x_{t_i}$, due to the multi-modality in $\nu_{t_i}$. The ALIs then either need to largely deviate from the reference trajectory (via a small regularizing weight) while sacrificing straightness, or comply with the linear reference trajectory, potentially interpolating through low-probability regions.

By employing a multi-marginal OT coupling and a piecewise-linear reference trajectory, however, the GAN's target distribution instead became less complex: with an OT-coupled triple $(x_0, x_{t_i}, x_1) \sim \pi(x_0, x_{t_i}, x_1)$, such as the one used in Rohbeck et al. [65], the adversarial objective rewards interpolants similar to intermediate marginal samples with small OT distances to $(x_0, x_1)$, and so the target distribution is less complicated.

This modification resulted in more accurate predictions and stable training for ALI-CFM, which comfortably outperformed the other methods. We argue that the superiority of ALI-CFM on this experiment is due to that the geometry varying across space ("time"), The 3D ST experiment thus provides a stringent stress test for the other methods, and serves as an empirical validation of the central thesis of this chapter: matching intermediate marginals in distribution, rather than enforcing pointwise interpolation, can be greatly beneficial in complex biological settings.

### 5.4.2 The challenge with multi-modality for multi-marginal flow matching methods

Above we showed how considering a multi-marginal OT coupling and more detailed reference interpolants could help to tackle multi-modality challenges, but at the expense of a more costly coupling and reference trajectory. Moreover, although the sensibility to multi-modality was decreased, the target distribution remains potentially multi-modal, especially in the 3D ST data setting.

Assume that the pair of end-marginal samples $(x_0, x_1)$ represent a pair of tumor annotated coordinates belonging to a single tumor volume with a loop as in Figure 5.4.2, which depicts a synthetic tumor geometry inspired by experimental findings in Mo et al. [9]. Let $x_{t_i}|x_0, x_1$ in the figure be the samples with non-negligible probability mass according to the OT-coupling $\pi(x_{t_i}|x_0, x_1)$, then the OT-coupling is multi-modal, which highlights a challenge for any MMFM method: for a given pair of $(x_0, x_1)$ samples and a multi-modal $\pi(x_{t_i}|x_0, x_1)$, which mode should the vector field integrate through? In the extreme case depicted in Figure 5.4.3, i.e. $x_0 = x_1 = 0$, $t_i = 0.5$, $x_{0.5} \in \{-1, 1\}$, and

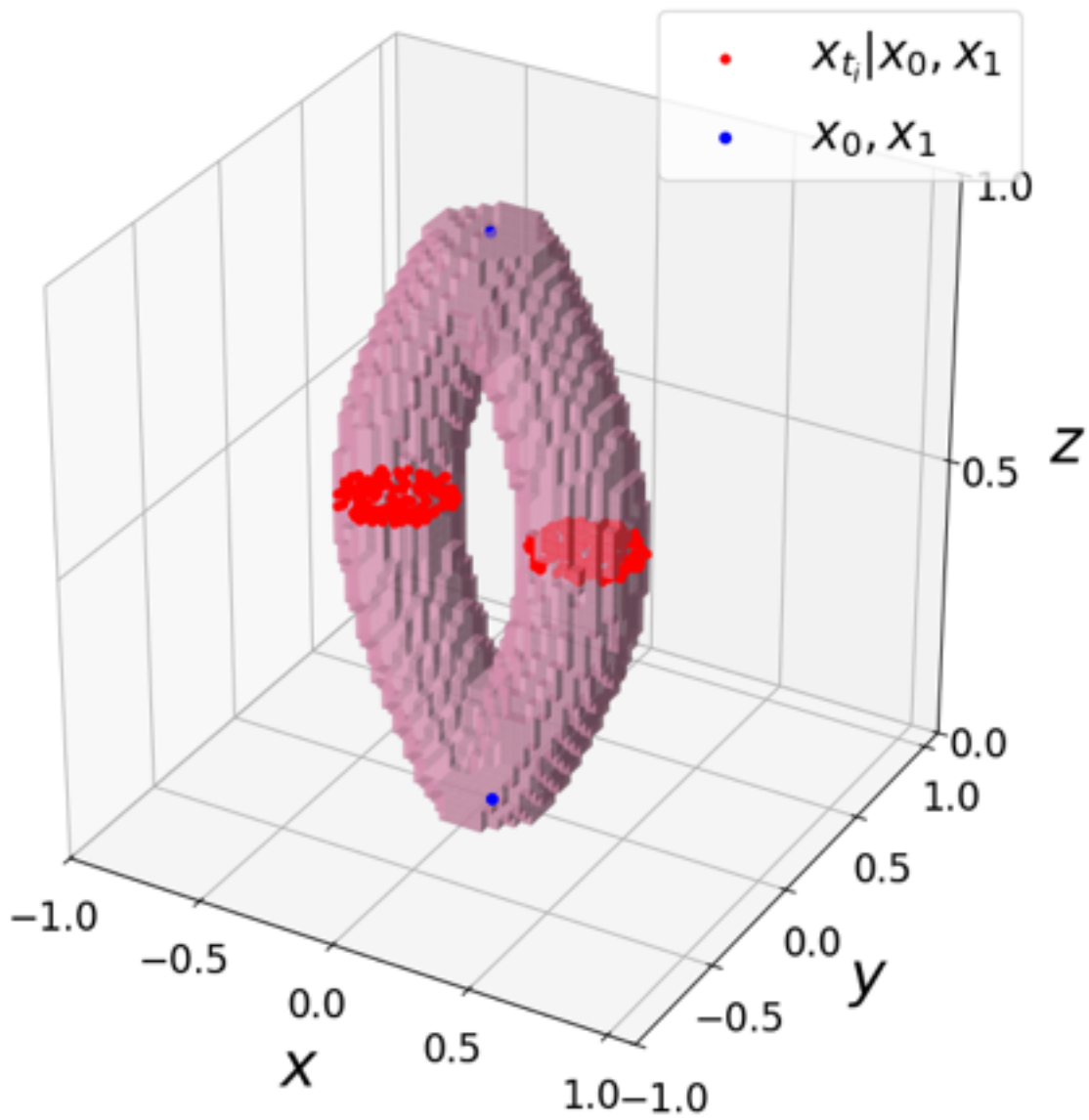

**Figure 5.4.2:** A synthetic geometry of a tumor volume, inspired by the experimental findings in Mo et al. [9, Fig. 4].

$\pi(x_{0.5} = 1|x_0, x_1) = \pi(x_{0.5} = -1|x_0, x_1) = 0.5$, the vector field will be fitted to

$$\frac{dx_t}{dt} = \begin{cases} 2, & t \in (0, 0.5) \\ -2, & t \in (0.5, 1) \end{cases} \tag{5.4.23}$$

in half of the training iterations, and in the other half of the training iterations it will be fitted to

$$\frac{dx_t}{dt} = \begin{cases} -2, & t \in (0, 0.5) \\ 2, & t \in (0.5, 1) \end{cases} \tag{5.4.24}$$

resulting in a vector field that, when initialized in $x_0 = 0$, deterministically produces an integrated trajectory constantly close to $0$, i.e. $x_t = 0$ for all $t \in [0, 1]$.

In the next section, we show how to leverage the variational mixture literature to construct multi-modal interpolants, learned via a scheme akin to the one proposed in Kviman et al. [4]—the adversarial objective is replaced with a MISELBO-based objective.

## 5.5 Multi-marginal flow matching with mixtures of variational interpolants

The setting is again interpolant learning in MMFM, but instead of using a GAN to match the intermediate marginal distributions, we instead infer variational mixture distributions that directly approximate the intermediate marginals using a multi-marginal formulation of MISELBO (see Equation (4.2.7)). The method described in this section is a product of Paper A-D, and is a novel contribution to the MMFM field.

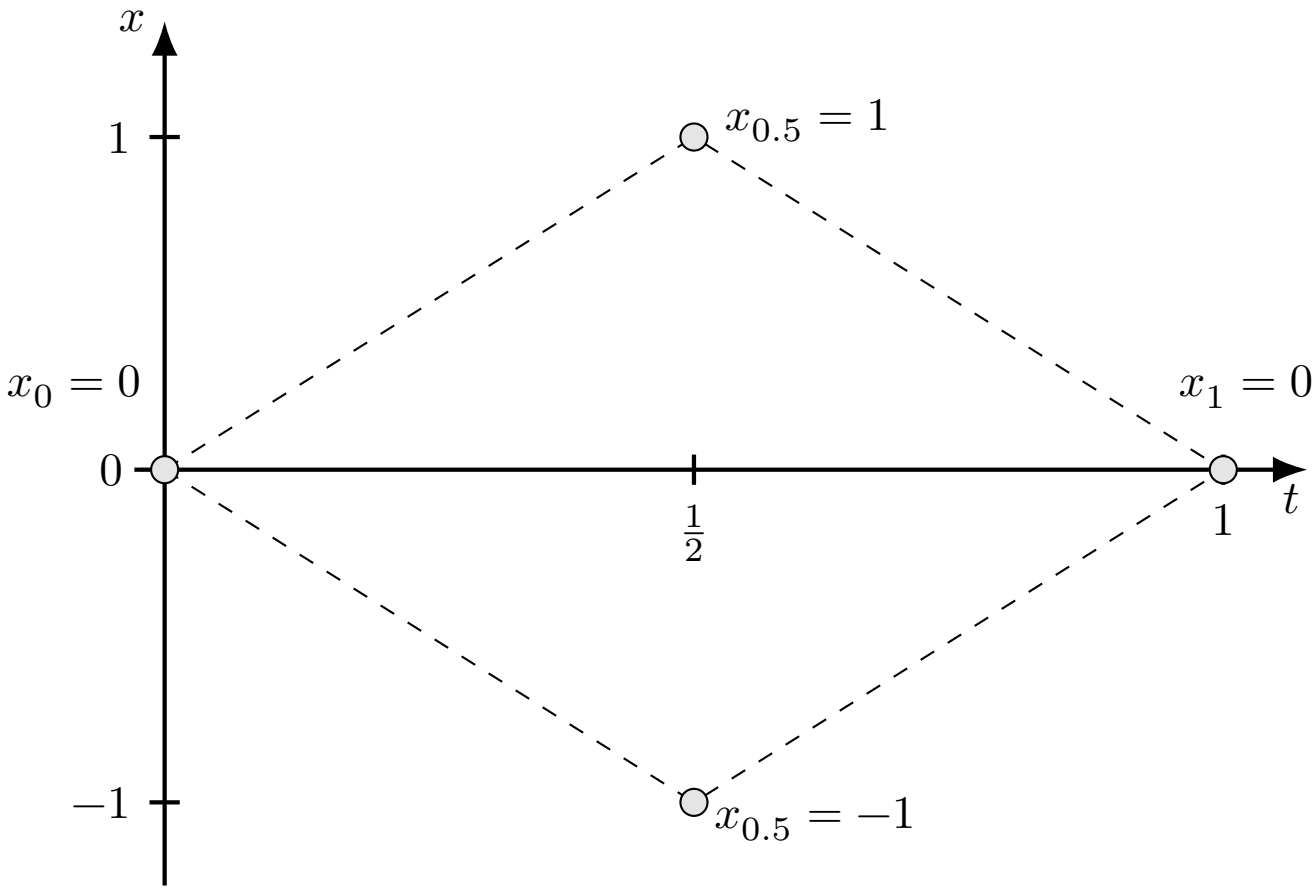

**Figure 5.4.3:** Toy multi-marginal setting: $x_0 = 0$, $x_1 = 0$ and $\pi(x_{0.5}|x_0, x_1) = \frac{1}{2}\delta_{-1} + \frac{1}{2}\delta_1$. Dashed lines indicate the possible piecewise linear interpolants.

### 5.5.1 Target marginal densities

As in the two previous sections, we assume access to $K$ marginal datasets $\mathcal{D}_{t_1:t_K}$, where $0 = t_1 < ... < t_K = 1$ and each $x_{t_i} \in \mathcal{D}_{t_i}$ is a draw from the marginal distribution $\nu_{t_i}$. Below we will formulate the two terms in the target marginal densities constructed in the marginals $k = 2, ..., K-1$. As in Kviman et al. [4], we do not approximate the end-marginal distributions, $\nu_0$ and $\nu_1$.

**Prior distributions** Now, conditioned on a draw of end-marginal pairs from a coupling $\pi(x_0, x_1)$, we consider prior distributions in each marginal time stamp denoted $p_{t_i}(x_{t_i}|x_0, x_1)$. The choice of parametric form of the prior is unconstrained, but here we choose it to be Gaussian with mean in the linear reference trajectory in Equation (5.4.19), $\ell(x_0, x_1, t_i) = t_i x_1 + (1 - t_i)x_0$, and some fixed variance $\sigma_0^2$:

$$p_{t_i}(x_{t_i}|x_0, x_1) = \mathcal{N}(\ell(x_0, x_1, t_i), \sigma_0^2). \tag{5.5.25}$$

Note the connection between the prior distribution and the regularizing term in the ALI objective (i.e., Equation (5.4.20)). The prior acts as a regularizer that penalizes deviations from the chosen reference trajectory. Specifically, for the Gaussian choice with the linear reference, the prior acts as a regularizing term with regularizing weight proportional to $1/\sigma_0^2$, which we show more formally below (see Section 5.5.3).

**Likelihood functions** To make the MISELBO objective well-defined for continuous variational approximations, we construct a new Gaussian mixture model likelihood function by placing Gaussian kernels with variance $s^2$ centered in marginal samples, i.e.

$$p_{t_i}(\mathcal{D}_{t_i}|x_{t_i}) = \frac{1}{|\mathcal{D}_{t_i}|} \sum_{x'_{t_i} \in \mathcal{D}_{t_i}} \mathcal{N}(x_{t_i}|x'_{t_i}, s^2), \tag{5.5.26}$$

where $|\cdot|$ denotes the cardinality operator. As such, Equation (5.5.26) can be seen as an approximation of the $\nu_{t_i}$ marginals using a continuous function based on $\mathcal{D}_{t_i}$.

In practice, if the dataset is too large, the mixture components in the likelihood can instead be centered around a subset (minibatch) of the marginal samples. Note that Equation (5.5.26) can also be viewed as the average likelihood of the marginal samples w.r.t. a Gaussian centered in $x_{t_i}$, which is equivalent to the mixture definition due to the symmetry of the Gaussian pdf.

As for the choice of the prior distribution, the likelihood function can be constructed in arbitrarily many ways. However, when using a mixture model, it is sensible to choose one that sums over the components in contrast to a product of components, as the latter model would result in a posterior that concentrates as we observe more marginal samples. Instead, the formulation in Equation (5.5.26) induces a multi-modal posterior with higher capacity to accurately model the complexity of the marginal distribution, $\nu_{t_i}$.

### 5.5.2 Mixture of variational interpolants

To approximate the target marginal densities introduced above, we will consider a time-dependent variational mixture distribution with $A$ equally weighted Gaussian components, conditioned on end-marginal samples,

$$q_t(x|x_0, x_1) = \frac{1}{A} \sum_{a=1}^{A} q_t^a(x|x_0, x_1), \tag{5.5.27}$$

where each mixture component has a mean which we parameterize as

$$\mu_t^a(x_0, x_1) = (1-t)x_0 + tx_1 + t(1-t)f_{\mu_a}(x_0, x_1, t), \tag{5.5.28}$$

where $f_{\mu_a}$ is a neural network as in Equation (5.4.16). We model the standard deviation of each component using a neural network scaled by a time-dependent coefficient,

$$\sigma_t^a = \sqrt{t(1-t)f_{\sigma_a}(x_0, x_1, t)}, \tag{5.5.29}$$

which ensures that

$$\lim_{t\to 0} q_t(x|x_0, x_1) = \delta_{x_0} \tag{5.5.30}$$

$$\lim_{t\to 1} q_t(x|x_0, x_1) = \delta_{x_1}, \tag{5.5.31}$$

and so the variational interpolants follow the correct end-marginal distributions (assume an independent coupling $\pi(x_0, x_1) = \nu_0\nu_1$ without loss of generalization to an OT coupling [62])

$$\int_{(x_0,x_1)} q_0(x|x_0, x_1)d\pi(x_0, x_1) = \int_{x_0} \delta_{x_0} d\nu_0(x_0) = \nu_0, \tag{5.5.32}$$

and vice versa for $t = 1$.

Indeed, the parameterization in Equation (5.5.28) is almost identical to Equation (5.4.16), with a notable difference that, instead of directly modeling the interpolants, Equation (5.5.28) parameterizes the distribution of the interpolants (i.e., the GMM component means).

### 5.5.3 The multi-marginal MISELBO objective

For a given $t_i \sim U\{t_2, ..., t_{K-1}\}$ and a $(x_0, x_1)$ pair, we now formulate the conditional MISELBO objective for multi-marginal interpolant learning,

$$\mathcal{L}_{\text{MIS}}(t_i, x_0, x_1) = \frac{1}{A}\sum_{a=1}^{A} \mathbb{E}_{x^a_{t_i} \sim q^a_{t_i}(x|x_0,x_1)} \left[ \log \frac{p_{t_i}(\mathcal{D}_{t_i}|x^a_{t_i}) p_{t_i}(x^a_{t_i}|x_0, x_1)}{\frac{1}{A}\sum_{a'=1}^{A} q^{a'}_{t_i}(x^a_{t_i}|x_0, x_1)} \right], \tag{5.5.33}$$

or, more compactly, as

$$\mathcal{L}_{\text{MIS}}(t_i, x_0, x_1) = \mathbb{E}_{x_{t_i} \sim q_{t_i}(x|x_0,x_1)} \left[ \log \frac{p_{t_i}(\mathcal{D}_{t_i}|x_{t_i}) p_{t_i}(x_{t_i}|x_0, x_1)}{q_{t_i}(x_{t_i}|x_0, x_1)} \right]. \tag{5.5.34}$$

Neatly, we can use the different MISELBO estimators established in Hotti et al. [3] to efficiently optimize this objective. Moreover, it is straightforward (see Equation (6.0.1)) to show that

$$\mathbb{E}_{(x_0,x_1)\sim\pi(x_0,x_1)}[\mathcal{L}_{\text{MIS}}(t_i, x_0, x_1)] \leq \log p_{t_i}(\mathcal{D}_{t_i}). \tag{5.5.35}$$

However, more generally, if we also average over all $t_i$, we get the multi-marginal MISELBO objective

$$\bar{\mathcal{L}}_{\text{MIS}} = \mathbb{E}_{t_i\sim U\{t_2,\ldots,t_{K-1}\},(x_0,x_1)\sim\pi(x_0,x_1)} \left[\mathcal{L}_{\text{MIS}}(x_0, x_1, t_i)\right], \tag{5.5.36}$$

which admits a lower bound on a multi-marginal log-likelihood (derivation in Equation (6.0.1))

$$\bar{\mathcal{L}}_{\text{MIS}} \leq \log \frac{1}{K-2} \sum_{i=2}^{K-1} p_{t_i}(\mathcal{D}_{t_i}). \tag{5.5.37}$$

**Optimization perspectives** We maximize $\bar{\mathcal{L}}_{\text{MIS}}$ to infer the parameters of the mixture of variational interpolants, which will result in a mixture distribution with support on $\mathcal{X} \times [0, 1]$ that minimizes a multi-marginal KL posterior matching objective: formally, let $\phi$ denote the parameters of the variational mixture and write the expectation subscripts more compactly for legibility, then

$$\max_{\phi} \bar{\mathcal{L}}_{\text{MIS}} = \min_{\phi} \mathbb{E}_{t_i,(x_0,x_1)} \left[ \text{KL}\left( q_{t_i}(x|x_0,x_1) \| p_{t_i}(x|\mathcal{D}_{t_i}, x_0, x_1) \right) \right]. \tag{5.5.38}$$

The derivation of this objective is provided in Equation (6.0.8) in the Appendix.

We can gain an alternative, arguably more informative perspective by decomposing $\bar{\mathcal{L}}_{\text{MIS}}$ into two terms (derivation in Chapter 6),

$$\bar{\mathcal{L}}_{\text{MIS}} = -\mathbb{E}_{t_i,(x_0,x_1)} \left[ \text{KL}(q_{t_i} \| p_{t_i}) \right] \tag{5.5.39}$$

$$- \mathbb{E}_{t_i,(x_0,x_1)} \left[ \mathbb{H}(q_{t_i}, p_{t_i})) \right], \tag{5.5.40}$$

where (see Equation (6.0.18))

$$\mathbb{H}(q_{t_i}, p_{t_i}) = \frac{1}{2\sigma_0^2} \frac{1}{A} \sum_{a=1}^{A} \left( (\sigma_{t_i}^a - 0)^2 + \left( \mu_{t_i}^a - \ell(x_0, x_1, t_i) \right)^2 \right). \tag{5.5.41}$$

Minimizing the cross-entropy thus rewards small component variances (the first term) and regularizes deviations between the component means and the reference trajectory (second term).

Hence, maximizing $\bar{\mathcal{L}}_{\text{MIS}}$ from the view of Equation (5.5.39) amounts to fitting the mixture of variational interpolants to the target marginal densities in terms of KL, subject to the cross-entropy regularization constraint with regularization weight $1/2\sigma_0^2$.

Fortunately, the S2S and S2A estimators introduced in Section 4.4.1 can easily be applied to the multi-marginal MISELBO objective. Since the likelihood function in Equation (5.5.26) can be extremely multi-modal with as many modes as there are marginal samples, the S2S estimator should be particularly powerful to employ here (see Section 4.4.3 for experimental motivations).

**A note on uniqueness** The $L_2$-norm regularization from the cross-entropy term above makes it tempting to try to apply Theorem 3 to this particular setup, too. However, optimization with mixtures typically yields non-convex problems, and challenges in terms of identifiability. We do not rule out the possibility of guaranteeing uniqueness w.r.t. the learned probability paths induced by the mixture, but leave the exploration of such guarantees for future work.

Nonetheless, in the spirit of highlighting the red thread among the four contributing papers in this thesis, I predict that if $A = 1$, i.e. the variational interpolant distribution is a single Gaussian, and if we consider a fixed variational variance, then there should exist a $\mu_t^a(x_0, x_1)$ in Equation (5.5.28) that uniquely minimizes $\bar{\mathcal{L}}_{\text{MIS}}$. This proof and the corresponding formalization of the statement hinges on the closed-form expression between the Gaussian prior and Gaussian variational distribution.

### 5.5.4 MVI-CFM

Once our mixture of variational interpolants (MVIs) have been learned, we want to use them in order to train a vector field. Unfortunately, we cannot simply plug-in the MVI-based conditional flow into the CFM objective in Equation (5.2.8): without informing the learnable vector field about from which mixture component the sampled interpolant originated from, the resulting vector field would average out the mixture information and produce meaningless trajectories. Returning to Figure 5.4.3, ignoring the component labeling information would correspond to learning a vector field that predicts $x_t = 0$ for all $t \in [0, 1]$.

In Rohbeck et al. [65], the authors considered scenarios where the data might have been sampled within different contexts, such as various drug response experiments. To capture this contextual affect on the data dynamics, they proposed to condition the vector field with pre-defined labels of the origin of the data.

We will use this conditioning mechanism to make the vector field component-aware by taking mixture component labels $a$ as input, yielding the MVI-CFM loss

$$\mathcal{L}_{\text{MVI-CFM}} = \mathbb{E}_{t,x_t,a}\left[\left\|u_t^\theta(x_t, a) - \frac{dx_t}{dt}\right\|^2\right], \tag{5.5.42}$$

where

$$\begin{aligned} t &\sim U[0,1], && (5.5.43)\\ x_t &= \mu_t^a(x_0, x_1) + \sigma_t^a(x_0, x_1)\,\epsilon, && (5.5.44)\\ a &\sim U\{1, \ldots, A\} && (5.5.45)\\ \epsilon &\sim \mathcal{N}(0, I_n) && (5.5.46)\\ (x_0, x_1) &\sim \pi(x_0, x_1). && (5.5.47) \end{aligned}$$

This results in $A$ ODEs which we can independently simulate and collect trajectories from. Finally, we note that, due to the construction of our MVIs such that their inputs are merely end-marginal pair samples—apart from the time variable—Equation (5.5.42) does not require a multi-marginal OT coupling to do MMFM-based inference, in contrast to the comparable C-MMFM loss in Rohbeck et al. [65].

### 5.5.5 Future work

As explained in the start of this chapter, them main purpose for including a derivation of this novel interpolation method in the kappa was to cement the connection between the concepts, methods and ideas proposed in the four contributed paper [1, 2, 3, 4]. As such, experimentation and new theoretical results reside outside of the scope of the thesis. Nonetheless, multi-marginal flow matching with variational mixture interpolants is not merely a theoretical proof-of-concept or a thought experiment, but will surely prove useful in practice, too. Demonstrating its practical usefulness to tackle multi-modality challenges in multi-marginal flow matching problems is thus an important future work direction, and there are more insights to be extracted from the contributed papers and in this kappa to tap its full potential.

# 6 Conclusions and Future Work

In this kappa, I have provided a methodological walk-through from the start of variational inference development to flow matching, a modern generative artificial intelligence technique. Along this path, I have pointed out where my four contributed papers [1, 2, 3, 4] sit, and described their technical contributions.

Specifically, the first three papers provide guidance on how to easily move from over-simplified variational approximations to multi-modal alternative using ensembles [1], a learning objective for inferring variational mixture distributions [2], and how to make variational mixture inference more efficient [3]. To build intuition for the introduced concepts and methods, I proposed a new probability distribution (see Section 3.2), which served as the running target-distribution example for the experiments throughout Chapter 4.

In the fourth paper [4], we show how to avoid inaccurate dynamics modeling in multi-marginal flow matching by reformulating the conditional flows constraints: Instead of requesting interpolation through marginal samples pointwise, match the distribution of the interpolations with the marginal distribution. Under certain regularization constraints, this approach enjoys uniqueness guarantees.

Finally, motivated by practical challenges that I highlight in Section 5.4.1, I demonstrated how to combine the works in order to advance the field by deriving a new flow matching method purposed for tackling these challenges. Future work consists of transforming this derivation into a scientifically contributing method.

There are many statistical inference problems in computational biology where sophisticated, multi-modal approximate inference methods are needed. Not least in cancer, as discussed in Section 5.4.1. My wish is that the contributed works and the kappa in this thesis will help guide future development of 3D spatial transcriptomics modeling, and the field of generative artificial intelligence in general.

# Additional derivations for variational interpolants

**Upper bounding Equation (5.5.36) by the multi-marginal log-likelihood** Applying Jensen's inequality to Equation (5.5.36), we get

$$
\begin{aligned}
\bar{\mathcal{L}}_{\text{MIS}} &\leq \log \mathbb{E}_{t_i \sim U\{t_2,\dots,t_{K-1}\},(x_0,x_1)\sim\pi(x_0,x_1),x_{t_i}\sim q_{t_i}(x|x_0,x_1)} \left[ \frac{p_{t_i}(\mathcal{D}_{t_i}|x_{t_i}) p_{t_i}(x_{t_i}|x_0,x_1)}{q_{t_i}(x_{t_i}|x_0,x_1)} \right] \\
&= \log \mathbb{E}_{t_i \sim U\{t_2,\dots,t_{K-1}\},(x_0,x_1)\sim\pi(x_0,x_1)} \left[ p_{t_i}(\mathcal{D}_{t_i}|x_t) p_{t_i}(x_t|x_0,x_1) \right] && (6.0.1) \\
&= \log \mathbb{E}_{t_i \sim U\{t_2,\dots,t_{K-1}\}} \left[ p_{t_i}(\mathcal{D}_{t_i}|x_t) p_{t_i}(x_t) \right] && (6.0.2) \\
&= \log \frac{1}{K-2} \sum_{i=2}^{K-1} p_{t_i}(\mathcal{D}_{t_i}). && (6.0.3)
\end{aligned}
$$

**Maximizing $\bar{\mathcal{L}}_{\text{MIS}}$ is equivalent to minimizing an MMFM KL objective** In a couple of steps, using Bayes' rule and that $\max\ -f(x) = \min f(x)$, we can show that maximizing $\bar{\mathcal{L}}_{\text{MIS}}$ is equivalent to minimizing an MMFM KL objective:

$$
\begin{aligned}
\max_\phi \bar{\mathcal{L}}_{\text{MIS}} &= \max_\phi \mathbb{E}_{t_i,(x_0,x_1),x_{t_i}|(x_0,x_1)} \left[ \log \frac{p_{t_i}(\mathcal{D}_{t_i}|x_{t_i}) p_{t_i}(x_{t_i}|x_0,x_1)}{q_{t_i}(x_{t_i}|x_0,x_1)} \right] && (6.0.4) \\
&= \max_\phi \mathbb{E}_{t_i,(x_0,x_1),x_{t_i}|(x_0,x_1)} \left[ \log \frac{p_{t_i}(x_{t_i}|\mathcal{D}_{t_i},x_0,x_1)}{q_{t_i}(x_{t_i}|x_0,x_1)} \right] + && (6.0.5) \\
&\quad \max_\phi\ - \mathbb{E}_{t_i,(x_0,x_1)} \left[ \log p_{t_i}(\mathcal{D}_{t_i},x_0,x_1) \right] && (6.0.6) \\
&= \max_\phi\ - \mathbb{E}_{t_i,(x_0,x_1),x_{t_i}|(x_0,x_1)} \left[ \log \frac{q_{t_i}(x_{t_i}|x_0,x_1)}{p_{t_i}(x_{t_i}|\mathcal{D}_{t_i},x_0,x_1)} \right] && (6.0.7) \\
&= \min_\phi \mathbb{E}_{t_i,(x_0,x_1)} \left[ \text{KL}\left( q_{t_i}(x|x_0,x_1) \| p_{t_i}(x|\mathcal{D}_{t_i},x_0,x_1) \right) \right]. && (6.0.8)
\end{aligned}
$$

**Derivation of closed-form cross-entropy and decomposition of the multi-marginal MISELBO objective** decomposing Equation (5.5.34) into two terms,

$$\mathcal{L}_{\text{MIS}}(t_i, x_0, x_1) = \underbrace{\mathbb{E}_{x_{t_i} \sim q_{t_i}(x|x_0, x_1)} \left[ \log \frac{p_{t_i}(\mathcal{D}_{t_i}|x_{t_i})}{q_{t_i}(x_{t_i}|x_0, x_1)} \right]}_{-\text{KL}(q_{t_i} \| p_{t_i})} \tag{6.0.9}$$

$$+ \underbrace{\mathbb{E}_{x_{t_i} \sim q_{t_i}(x|x_0, x_1)} \left[ \log p_{t_i}(x_{t_i}|x_0, x_1) \right]}_{-\mathbb{H}(q_{t_i}, p_{t_i})}, \tag{6.0.10}$$

i.e. the negative KL divergence from the variational mixture to the likelihood function[1] and the negative cross-entropy between the variational mixture and the prior.

Analyzing the cross-entropy, it turns out that $\mathbb{H}(q_{t_i}, p_{t_i})$ has a closed form expressed in terms of the $L_2$-norm since the variational mixture is assumed to be a GMM, and since mixture expectations are linear

$$\mathbb{H}(q_{t_i}, p_{t_i}) = \frac{1}{A} \sum_{a=1}^{A} \mathbb{H}(q_{t_i}^a, p_{t_i}). \tag{6.0.11}$$

Each summand in Equation (6.0.11) is a cross-entropy operation between two Gaussians, which has a well-known closed form

$$\begin{aligned}
\mathbb{H}(q_{t_i}^a, p_{t_i}) &= -\mathbb{E}_{q_{t_i}^a(x|x_0,x_1)}[\log p_{t_i}(x|x_0, x_1)] && (6.0.12)\\
&= -\mathbb{E}_{q_{t_i}^a(x|x_0,x_1)} \left[ \log \frac{1}{\sqrt{2\pi}\sigma_0} - \frac{1}{2\sigma_0^2}(x - \ell(x_0, x_1, t_i))^2 \right] && (6.0.13)\\
&= \log \sqrt{2\pi}\sigma_0 + \frac{1}{2\sigma_0^2} \mathbb{E}_{q_{t_i}^a(x|x_0,x_1)} \left[ (x - \ell(x_0, x_1, t_i))^2 \right] && (6.0.14)\\
&\propto \frac{1}{2\sigma_0^2} \left( \mathbb{E}_{q_{t_i}^a(x|x_0,x_1)} \left[ x^2 \right] - 2\ell(x_0, x_1, t_i) \mathbb{E}_{q_{t_i}^a(x|x_0,x_1)}[x] + \ell(x_0, x_1, t_i)^2 \right) && (6.0.15)\\
&= \frac{(\sigma_{t_i}^a)^2}{2\sigma_0^2} + \frac{1}{2\sigma_0^2} \left( \mu_{t_i}^a - \ell(x_0, x_1, t_i) \right)^2, && (6.0.16)
\end{aligned}$$

and so the GMM-to-Gaussian cross-entropy writes (proportionally)

$$\begin{aligned}
\mathbb{H}(q_{t_i}, p_{t_i}) &\propto \frac{1}{A} \sum_{a=1}^{A} \frac{(\sigma_{t_i}^a)^2}{2\sigma_0^2} + \frac{1}{A} \sum_{a=1}^{A} \frac{1}{2\sigma_0^2} \left( \mu_{t_i}^a - \ell(x_0, x_1, t_i) \right)^2 && (6.0.17)\\
&= \frac{1}{2\sigma_0^2} \frac{1}{A} \sum_{a=1}^{A} \left( (\sigma_{t_i}^a - 0)^2 + \left( \mu_{t_i}^a - \ell(x_0, x_1, t_i) \right)^2 \right). && (6.0.18)
\end{aligned}$$

[1]The likelihood function is chosen as a GMM in Equation (5.5.26) and so it is a normalized density and the term is indeed a well-defined KL divergence.

Apply the time- and coupling-expectations to Equation (6.0.9),

$$\mathbb{E}_{t_i,(x_0,x_1)}\left[\mathcal{L}_{\text{MIS}}(t_i,x_0,x_1)\right] = \bar{\mathcal{L}}_{\text{MIS}} = -\,\mathbb{E}_{t_i(x_0,x_1)}\left[\text{KL}(q_{t_i}\|p_{t_i})\right] \tag{6.0.19}$$

$$-\,\mathbb{E}_{t_i,(x_0,x_1)}\left[\mathbb{H}(q_{t_i},p_{t_i}))\right]. \tag{6.0.20}$$